\documentclass[preprint]{article}

     \PassOptionsToPackage{numbers, compress}{natbib}

\usepackage[preprint]{sty/neurips_2020}

\usepackage{sty/macros_neurips}

\usepackage{hyperref}       
\definecolor{blued}{RGB}{70,197,221}
\definecolor{pearOne}{HTML}{2C3E50}
\definecolor{pearTwo}{HTML}{A9CF54}
\definecolor{pearTwoT}{HTML}{C2895B}
\definecolor{pearThree}{HTML}{FF69B4}
\definecolor{pearThreeAndHalf}{HTML}{FF6969}
\colorlet{titleTh}{pearOne}
\colorlet{bull}{pearTwo}
\definecolor{pearcomp}{HTML}{B97E29}
\definecolor{pearFour}{HTML}{588F27}
\definecolor{pearFith}{HTML}{ECF0F1}
\definecolor{pearDark}{HTML}{2980B9}
\definecolor{pearDarker}{HTML}{F330DB}
\hypersetup{
	colorlinks,
	citecolor=pearDark,
	linkcolor=pearThreeAndHalf,
	urlcolor=pearDarker}

\usepackage{cleveref}

\usepackage{todonotes}

\title{Planning in Markov Decision Processes with Gap-Dependent Sample Complexity}

\author{%
	Anders Jonsson \\
	Universitat Pompeu Fabra \\
	\texttt{anders.jonsson@upf.edu} \\
	\And
	Emilie Kaufmann \\
	CNRS \& ULille (CRIStAL), Inria SequeL  \\
	\texttt{emilie.kaufmann@univ-lille.fr} \\
	\And
	Pierre M\'enard \\
	Inria Lille, SequeL team \\
	\texttt{pierre.menard@inria.fr} \\
	\And
	Omar Darwiche Domingues \\
    Inria Lille, SequeL team \\
    \texttt{omar.darwiche-domingues@inria.fr} \\
    \And
    \hspace*{-5em} Edouard Leurent \\
    \hspace*{-5em} Renault \& Inria Lille, SequeL team \\
    \hspace*{-5em} \texttt{edouard.leurent@inria.fr} \\
    \And
    \hspace*{-2em} Michal Valko \\
    \hspace*{-2em} DeepMind Paris \\
    \hspace*{-2em} \texttt{valkom@deepmind.com}
}

\begin{document}

\maketitle

\begin{abstract}%
  We propose \OurAlgorithm{}, a new trajectory-based Monte-Carlo Tree Search algorithm for planning in a Markov Decision Process in which transitions have a finite support. We prove an upper bound on the number of calls to the generative models needed for \OurAlgorithm{} to identify a near-optimal action with high probability. This problem-dependent \emph{sample complexity} result is expressed in terms of the \emph{sub-optimality gaps} of the state-action pairs that are visited during exploration. Our experiments reveal that \OurAlgorithm{} is also effective in practice, in contrast with other algorithms with sample complexity guarantees in the fixed-confidence setting, that are mostly theoretical.
\end{abstract}

\documentclass[../neurips2020.tex]{subfiles}

\section{Introduction}
In reinforcement learning (RL), an agent repeatedly takes \emph{actions} and observes \emph{rewards} in an unknown environment described by a \emph{state}. Formally, the environment is a Markov Decision Process (MDP) $\mathcal{M}=\tuple{\mathcal{S},\mathcal{A},p,r}$, where $\cS$ is the state space, $\cA$ the action space, $p = \cpa{p_h}_{h \geq 1}$ a set of transition kernels and $r = \cpa{r_h}_{h\geq 1}$ a set of reward functions. By taking action $a$ in state $s$ at step $h$, the agent reaches a state $s'$ with probability $p_h(s'|s, a)$ and receives a random reward with mean $r_h(s, a)$. A common goal is to learn a policy $\pi= (\pi_h)_{h \geq 1}$ that maximizes cumulative reward by taking action $\pi_h(s)$ in state $s$ at step $h$. If the agent has access to a generative model, it may \emph{plan} before acting by generating additional samples in order to improve its estimate of the best action to take next.

In this work, we consider \emph{Monte-Carlo planning} as the task of recommending a good action to be taken by the agent in a given state $s_1$, by using samples gathered from a generative model. Let $Q^\star(s_1, a)$ be the maximum cumulative reward, in expectation, that can be obtained from state $s_1$ by first taking action $a$, and let $\hat{a}_n$ be the recommended action after $n$ calls to the generative model. The quality of the action recommendation is measured by its \emph{simple regret}, defined as
$
	\bar{r}_n(\hat{a}_n) \eqdef V^\star(s_1) - Q^\star(s, \hat{a}_n), \; \mbox{where} \; V^\star(s_1) \eqdef  \max_a Q^\star(s_1, a)$.

We propose an algorithm in the \emph{fixed confidence} setting $(\epsilon, \delta)$: after $n$ calls to the generative model, the algorithm should return an action $\hat{a}_n$ such that $\bar{r}_n(\hat{a}_n) \leq \epsilon$ with probability at least $1-\delta$. We prove that its \emph{sample complexity} $n$ is bounded in high probability by a quantity that depends on the sub-optimality gaps of the actions that are applicable in state $s_1$. We also provide experiments showing its effectiveness. The only assumption that we make on the MDP is that the support of the transition probabilities $p_h(\cdot|s, a)$ should have cardinality bounded by $B < \infty$, for all $s$, $a$ and $h$.

Monte-Carlo Tree Search (MCTS) is a form of Monte-Carlo planning that uses a {\em forward model} to sample transitions from the current state, as opposed to a full generative model that can sample anywhere. Most MCTS algorithms sample {\em trajectories} from the current state~\cite{SurveyMCTS12}, and are widely used in \emph{deterministic} games such as Go. The AlphaZero algorithm~\cite{AlphaZero} guides planning using value and policy estimates to generate trajectories that improve these estimates. The MuZero algorithm~\cite{MuZero} combines MCTS with a model-based method which has proven useful for \emph{stochastic} environments. Hence \emph{efficient Monte-Carlo planning} may be instrumental for learning better policies. Despite their empirical success, little is known about the sample complexity of state-of-the-art MCTS algorithms. 

\paragraph{Related work} The earliest MCTS algorithm with theoretical guarantees is Sparse Sampling~\cite{Kearns02SS}, whose sample complexity is polynomial in $1/\epsilon$ in the case $B<\infty$ (see Lemma~\ref{lemma:SS}).
However, it is not trajectory-based and does not select actions adaptively, making it very inefficient in practice.

Since then, adaptive planning algorithms with small sample complexities have been proposed in different settings with different optimality criteria. In Table~\ref{tab:different_settings_in_literature}, we summarize the most common settings, and in Table~\ref{tab:related_work_sample_complexity}, we show the sample complexity of related algorithms (omitting logarithmic terms and constants) when $B<\infty$. Algorithms are either designed for a discounted setting with $\gamma < 1$ or an episodic setting with  horizon $H$. Sample complexities are stated in terms of the accuracy $\epsilon$ (for algorithms with fixed-budget guarantees we solve $\bE\spa{\bar{r}_n} = \epsilon$ for $n$), the number of actions $K$, the horizon $H$ or the discount factor $\gamma$ and a problem-dependent quantity $\kappa$ which is a notion of branching factor of near-optimal nodes whose exact definition varies.

A first category of algorithms rely on optimistic planning~\cite{SurveyRemiMCTS}, and require additional assumptions: a deterministic MDP \citep{hren2008optimistic}, the \emph{open loop} setting \citep{bubeck2010open,leurent2019practical} in which policies are sequences of actions instead of state-action mappings (the two are equivalent in MDPs with deterministic transitions), or an MDP with known parameters \citep{busoniu2012optimistic}. For MDPs with stochastic and unknown transitions, polynomial sample complexities have been obtained for StOP \citep{STOP14}, TrailBlazer \citep{TrailBlazer16} and SmoothCruiser \citep{SmoothCruiser19} but the three algorithms suffer from numerical inefficiency, even for $B<\infty$. Indeed, StOP explicitly reasons about policies and storing them is very costly, while TrailBlazer and SmoothCruiser require a very large amount of recursive calls even for small MDPs.
We remark that popular MCTS algorithms such as UCT~\cite{Kocsis06UCT} are not $(\epsilon,\delta)$-correct and do not have provably small sample complexities.

In the setting $B<\infty$, BRUE~\cite{Feldman14BRUE} is a trajectory-based algorithm that is anytime and whose sample complexity depends on the smallest sub-optimality gap $\Delta \eqdef\min_{a\neq a^\star}\pa{V^\star(s_1) - Q^\star(s_1, a)}$. For planning in deterministic games, gap-dependent sample complexity bounds were previously provided in a fixed-confidence setting \citep{Huang17StructuredBAI,BAIMCTS17}. Our proposal, \OurAlgorithm{}, can be viewed as a non-trivial adaptation of the \UGapEMCTS algorithm~\cite{BAIMCTS17} to planning in MDPs. The defining property of \OurAlgorithm{} is that it uses a best arm identification algorithm, UGapE \citep{gabillon2012best}, to select the first action in a trajectory, and performs optimistic planning thereafter, which helps refining confidence intervals on the intermediate Q-values. Best arm identification tools have been previously used for planning in MDPs~\cite{Pepels14SimpleMCTS,Tolpin12SRMCTS} and UGapE also served as a building block for StOP~\citep{STOP14}. 

Finally, going beyond worse-case guarantees for RL is an active research direction, and in a different context gap-dependent bounds on the regret have recently been established for tabular MDPs~\cite{simchowitz2019gaps,zanette2019gaps}.

\begin{table}[t]
	\caption{Different settings of planning algorithms in the literature}
	\label{tab:different_settings_in_literature}

	\begin{center}
	\begin{small}
		\begin{tabular}{@{}llll@{}}
			\toprule
			Setting                                     & Input                         & Output             & Optimality criterion \\ \midrule
			(1) Fixed confidence (action-based)& $\epsilon, \delta$ & $\widehat{a}_n$ & $\bP\pa{ \bar{r}_n(\widehat{a}_n) \leq \epsilon } \geq 1 - \delta$   \\
			(2) Fixed confidence (value-based) & $\epsilon, \delta$  & $\widehat{V}(s_1)$   & $\bP\pa{ |\widehat{V}(s_1) - V^\star(s_1)| \leq \epsilon} \geq 1 - \delta$           \\
			(3) Fixed budget  &  $n$ (budget) &  $\widehat{a}_n$ &  $\bE\spa{\bar{r}_n(\widehat{a}_n)}$ decreasing in $n$ \\[2pt]
			(4) Anytime       &  - &  $\widehat{a}_n$ &  $\bE\spa{\bar{r}_n(\widehat{a}_n)}$ decreasing in $n$ \\
			\bottomrule
		\end{tabular}
	\end{small}
	\end{center}

\end{table}

\begin{table}[t]
	\caption{Algorithms with sample complexity guarantees}
	\label{tab:related_work_sample_complexity}

	\begin{center}
	\begin{small}
		\begin{tabular}{@{}l@{\hspace*{5pt}}l@{\hspace*{5pt}}l@{\hspace*{5pt}}l@{}}
			\toprule
			Algorithm         & Setting & Sample complexity & Remarks          \\ \midrule
			Sparse Sampling \citep{Kearns02SS}  & (1)-(2)   &  $H^5(BK)^H/\epsilon^2$  or $\epsilon^{-\left(2 + \frac{\log(K)}{\log(1/\gamma)}\right)}$              &                  proved in Lemma~\ref{lemma:SS}\\
			OLOP \citep{bubeck2010open}   & (3)     &   $\epsilon^{-\max\pa{2,  \frac{\log\kappa}{\log (1/\gamma)}}}$                & open loop, $\kappa \in [1, K]$   \\
			OP \citep{busoniu2012optimistic} & (4)     &   $\epsilon^{-\frac{\log\kappa}{\log (1/\gamma)}}$                 & known MDP, $\kappa \in [0, BK]$        \\[2pt]
			BRUE \citep{Feldman14BRUE}              & (4)     & ${H^4(BK)^H}/{\Delta^2}  $                 &    minimal gap $\Delta$               \\
			StOP \citep{STOP14}             & (1)     &  $\epsilon^{-\pa{2+\frac{\log\kappa}{\log (1/\gamma)}+o(1)}}  $                 &      $\kappa \in [0, BK]$            \\
			TrailBlazer \citep{TrailBlazer16}       & (2)     &  $\epsilon^{-\max\pa{2,\frac{\log(B\kappa)}{\log (1/\gamma)}+o(1)}} $                   & $\kappa \in [1, K]$                  \\[2pt]
			SmoothCruiser \citep{SmoothCruiser19}     & (2)     &  $\epsilon^{-4}$                 & only regularized MDPs \\
			\bottomrule
			\OurAlgorithm{} (ours) & (1)  & $\sum_{a_1\in\cA} \frac{H^2(BK)^{H-1}B}{\left(\Delta_1(s_1,a_1) \vee \Delta \vee  \epsilon \right)^2}$  & see Corollary \ref{cor:firstbound}\\
			\bottomrule
		\end{tabular}
	\end{small}
	\end{center}

\end{table}

\paragraph{Contributions} We present \OurAlgorithm{}, a new MCTS algorithm for planning in the setting $B<\infty$.  \OurAlgorithm{} performs efficient Monte-Carlo planning in the following sense: First, it is a simple trajectory-based algorithm which performs well in practice and only relies on a forward model. Second, while most practical MCTS algorithms are not well understood theoretically, we prove upper bounds on the sample complexity of \OurAlgorithm{}. Our bounds depend on the \emph{sub-optimality gaps} associated to the state-action pairs encountered during exploration. This is in contrast to StOP and TrailBlazer, two algorithms for the same setting, whose guarantees depend on a notion of near-optimal nodes which can be harder to interpret, and that can be inefficient in practice. In the anytime setting, BRUE also features a gap-dependent sample complexity, but only through the worst-case gap $\Delta$ defined above. As can be seen in Table~\ref{tab:different_settings_in_literature}, the upper bound for \OurAlgorithm{} given in Corollary~\ref{cor:firstbound} improves over that of \BRUE{} as it features the gap of each possible first action $a_1$, $\Delta_1(s_1,a_1) = V^\star(s_1) -Q_1^\star(s_1,a_1)$, and scales better with the planning horizon $H$. Furthermore, our proof technique relates the \emph{pseudo-counts} of any trajectory prefix to the gaps of state-action pairs on this trajectory, which evidences the fact that \OurAlgorithm{} does not explore trajectories uniformly.

\section{Learning Framework and Notation}\label{sec:setup}

We consider a \emph{discounted episodic setting} where $H \in \N^\star$ is a horizon and $\gamma \in (0,1]$ a discount parameter. The transition kernels $p=(p_1,\ldots,p_H)$ and reward functions $r=(r_1,\ldots,r_H)$ can have distinct definitions in each step of the episode.
The optimal value of selecting action $a$ in state $s_1$ is
\[Q^\star(s_1,a) = \max_{\pi}\bE^{\pi}\left[\left.\sum_{h=1}^H \gamma^{h-1}r_h(s_h,a_h)\right| a_1 = a \right],\]
where the supremum is taken over (deterministic) policies $\pi = (\pi_1,\dots,\pi_H)$, and the expectation is on a trajectory $s_1,a_1,\dots,s_h,a_h$ where  $s_{h} \sim p_{h-1}( \cdot | s_{h-1},a_{h-1})$ and $a_h = \pi_h(s_h)$ for $h\in [2,H]$. With this definition, an optimal action in state $s_1$ is
$a^\star \in \text{argmax}_{a \in \cA(s_1)} Q^\star(s_1,a)$.

We assume that there is a maximal number $K$ of actions available in each state, and that, for each $(s,a)$, the support of $p_h(\cdot |s,a)$ is bounded by $\support$: that is, $\support$ is the maximum number of possible next states when applying any action.
We further assume that the rewards are bounded in $[0,1]$.
 For each pair of integers $i,h$ such that $i\leq h$, we introduce the notation $[i,h]=\{i,\ldots,h\}$ and $[h]=[1,h]$.

\paragraph{$(\epsilon,\delta)$-correct planning} A sequential planning algorithm proceeds as follows. In each episode $t$, the agent uses a deterministic policy on the form $\pi^t=(\pi_1^t,\ldots,\pi_H^t)$ to generate a trajectory $(s_1,a_1^t,r_1^t,\ldots,s_H^t,a_H^t,r_H^t)$, where $a_h^t = \pi_h^t(s_h^t)$, $r_h^t$ is a reward with expectation $r_h(s_h^t,a_h^t)$ and $s_{h+1}^t \sim p_h(\cdot | s_h^t,a_h^t)$. After each episode the agent decides whether it should perform a new episode to refine its guess for a near-optimal action, or whether it can stop and make a guess. We denote by $\tau$ the stopping rule of the agent, that is the number of episodes performed, and $\hat{a}_\tau$ the guess.

We aim to build an $(\epsilon,\delta)$-correct algorithm, that is an algorithm that outputs a guess $\hat{a}_\tau$ satisfying
\begin{equation}\bP\left(Q^\star(s_1,\hat{a}_\tau) > Q^\star(s_1,a^\star) - \epsilon\right) \geq 1 - \delta \ \ \ \ \Leftrightarrow \ \ \ \ \bP\left(\bar{r}_{(H\tau)}\left(\hat{a}_\tau\right)\leq \epsilon\right)\geq 1 - \delta\label{target:PAC}\end{equation}
while using as few calls to the generative model $n = H\tau$ (i.e.~as few episodes $\tau$) as possible.

Our setup permits to propose algorithms for planning in the undiscounted episodic case (in which our bounds will not blow up when $\gamma=1$) and in discounted MDPs with infinite horizon. Indeed, choosing $H$ such that $2\gamma^H/(1-\gamma)\leq {\epsilon}$, an $(\epsilon,\delta)$-correct algorithm for the discounted episodic setting recommends an action that is $2\epsilon$-optimal for the discounted infinite horizon setting.

\paragraph{A (recursive) baseline} Sparse Sampling \citep{Kearns02SS} can be tuned to output a guess $\hat{a}$ that satisfies~\eqref{target:PAC}, as specified in the following lemma, which provides a baseline for our undiscounted episodic setting (see Appendix~\ref{proof:SS}). Note that Sparse Sampling is not strictly sequential as it does not repeatedly select trajectories.

\begin{lemma}\label{lemma:SS}
	If $B < \infty$, Sparse Sampling using horizon $H$ and performing $\BigO{({H^5}/{\epsilon^2})\log\pa{{BK}/{\delta}}}$ transitions in each node is $(\epsilon,\delta)$-correct with sample complexity $O(n_{\text{SS}})$ for $n_{\text{SS}} \eqdef {H^5(BK)^H}/{\epsilon^2}$.
\end{lemma}

\paragraph{Structure of the optimal Q-value function} In our algorithm, we will build estimates of the intermediate Q-values, that are useful to compute the optimal Q-value function $Q^\star(s_1,a)$. Defining
\[Q_h(s_h,a_h) = \max_{\pi} \bE^{\pi}\left[\left. \sum_{i=h}^H \gamma^{i-h} r(s_i,a_i) \right| s_h,a_h\right],\]
$Q^\star(s_1,a) = Q_1(s_1,a)$ and the optimal action-values $Q=(Q_1,\ldots,Q_H)$ can be computed recursively using the Bellman equations, where we use the convention $Q_{H+1}(\cdot,\cdot)=0$:\[Q_h(s_h,a_h) = r_h(s_h,a_h) + \gamma \sum_{s'} p_h(s'|s_h,a_h) \max_{a'} Q_{h+1}(s',a'), \;\; h\in[H].\]
Let $\pi^\star = (\pi_1^\star,\ldots,\pi_H^\star)$ denote a deterministic optimal policy where, for $h\in[H]$, $\pi_h^\star(s_h) = \arg\max_a Q_h(s_h,a)$, with ties arbitrarily broken. Hence the optimal value in $s_h$ is $Q_h(s_h,\pi_h^\star(s_h))$.

\documentclass[../neurips2020.tex]{subfiles}

\section{The \OurAlgorithm{} Algorithm}\label{sec:algorithm}

In this section we present \OurAlgorithm{}, a generalization of \UGapE~\citep{gabillon2012best} to Monte-Carlo planning. Like BAI-MCTS for games \citep{BAIMCTS17} a core component is the construction of confidence intervals on $Q_1(s_1,a)$. The construction below generalizes that of OP-MDP \citep{busoniu2012optimistic} for known transition probabilities.

\paragraph{Confidence bounds on the $Q$-values} Our algorithm maintains empirical estimates, superscripted with the episode $t$, of the transition kernels $p$ and expected rewards $r$, which are assumed unknown.

Let $n_h^t(s_h,a_h,s_{h+1}) \eqdef \sum_{s=1}^t \ind\left((s_h^s,a_h^s,s_{h+1}^s)=(s_h,a_h,s_{h+1})\right)$ be the number of observations of transition $(s_h,a_h,s_{h+1})$, and $R_h^t(s_h,a_h) \eqdef \sum_{s=1}^t r_h^s(s_h,a_h) \ind\left((s_h^s,a_h^s)=(s_h,a_h)\right)$ the sum of rewards obtained when selecting $a_h$ in $s_h$. We define the empirical transition probabilities $\hat{p}^t$ and expected rewards $\hat{r}^t$ as follows, for state-action pairs such that $n_h^t(s_h,a_h)\eqdef\sum_s n_h^t(s_h,a_h,s) > 0$:
\begin{align*}
\hat{p}_h^t(s_{h+1}|s_h,a_h) \eqdef \frac {n_h^t(s_h,a_h,s_{h+1})} {n_h^t(s_h,a_h)},\ \ \text{and} \ \ \ \hat{r}_h^t(s_h,a_h) \eqdef\frac {R_h^t(s_h,a_h)} {n_h^t(s_h,a_h)}.
\end{align*}

As rewards are bounded in $[0,1]$, we define the following Kullback-Leibler upper and lower confidence bounds on the mean rewards $r_h(s_h,a_h)$ \cite{KLUCBJournal}:
\begin{align*}
u_h^t(s_h,a_h) &\eqdef \max \left\{v : \kl\!\big(\hat{r}_h^t(s_h,a_h),v\big) \leq \frac {\beta^r(n_h^t(s_h,a_h), \delta)} {n_h^t(s_h,a_h)} \right\},\\
\ell_h^t(s_h,a_h) &\eqdef\min \left\{v : \kl\!\big(\hat{r}_h^t(s_h,a_h),v\big) \leq \frac {\beta^r(n_h^t(s_h,a_h), \delta)} {n_h^t(s_h,a_h)} \right\},
\end{align*}
where $\beta^r$ is an exploration function and $\kl(u,v)$ is the binary Kullback-Leibler divergence between two Bernoulli distributions $\Ber(u)$ and $\Ber(v)$: $\kl(u,v) = u \log \tfrac {u} {v} + (1-u) \log \tfrac {1-u} {1-v}$.
We adopt the convention that $u_h^t(s_h,a_h) =1, \ell_h^t(s_h,a_h) = 0$ when $n_h^t(s_h,a_h)=0$.

In order to define confidence bounds on the values $Q_h$, we introduce a confidence set on the probability vector $p_h(\cdot|s_h,a_h)$. We define  $\cC_h^t(s_h,a_h) = \Sigma_{B}$ if $n_h^t(s_h,a_h)=0$ and otherwise
\[\cC_h^t(s_h,a_h) \eqdef \left\{p\in \Sigma_{B} :  \KL\!\big(\hp_h^t(\cdot|s_h,a_h),p\big) \leq \frac {\beta^p(n_h^t(s_h,a_h), \delta)} {n_h^t(s_h,a_h)}\right\},\]
where $\Sigma_{B}$ is the set of probability distribution over $B$ elements, $\beta^p$ is an exploration function and $\KL(p,q)= \sum_{s \in \text{Supp}(p)}  p(s) \log \tfrac{ p(s)} {{q}(s)}$ is the Kullback-Leibler divergence between two categorical distributions $p$ and $q$ with supports satisfying $\text{Supp}(p) \subseteq \text{Supp}(q)$.

We now define our confidence bounds on the action values inductively. We use the convention $U_{H+1}^t(\cdot,\cdot)=L_{H+1}^t(\cdot,\cdot)=0$, and for all $h\in[H]$,
\begin{align*}
U_h^t(s_h,a_h) &= u_h^t(s_h,a_h) + \gamma \max_{p \in \cC_h^t(s_h,a_h)} \sum_{s'} p(s'|s_h,a_h) \max_{a'} U_{h+1}^t(s',a'), \\
L_h^t(s_h,a_h) &= \ell_h^t(s_h,a_h) + \gamma \min_{p \in \cC_h^t(s_h,a_h)} \sum_{s'} p(s'|s_h,a_h) \max_{a'} L_{h+1}^t(s',a').
\end{align*}
As explained in Appendix A of \cite{filippi2010optimism}, optimizing over these KL confidence sets can be reduced to a linear program with convex constraints, that can be solved efficiently with Newton Iteration, which has complexity $O(B\log(d))$ where $d$ is the desired digit precision. 

We provide in Section~\ref{subsec:correctness} an explicit choice for the exploration functions $\beta^r(n,\delta)$ and $\beta^p(n,\delta)$ that govern the size of the confidence intervals. Note that if the rewards or transitions are deterministic, or if we know $p$, we can adapt our confidence bounds by setting $\beta^p=0$ or $\beta^r=0$.

\paragraph{\OurAlgorithm} As any fixed-confidence algorithm, \OurAlgorithm{} depends on the tolerance parameter $\epsilon$ and the risk parameter $\delta$. The dependency in $\epsilon$ is explicit in the stopping rule  \eqref{def:Stopping}, while the dependency in $\delta$ is in the tuning of the confidence bounds, that depend on $\delta$.

After $t$ trajectories observed, \OurAlgorithm{} selects the $(t+1)$-st trajectory using the policy $\pi^{t+1}=(\pi^{t+1}_1,\dots,\pi^{t+1}_H)$ where the first action choice is made according to \UGapE: \[\pi^{t+1}_1(s_1) = \argmax{b \in \{b^t,c^t\}} \left[U_1^t(s_1,b) - L_1^t(s_1,b)\right]\,,\]
where $b^t$ is the current guess for the best action, which is the action $b$ with the smallest upper confidence bound on its gap $Q_1^\star(s_1,a^\star) - Q_1(s_1,b)$, and $c^t$ is some challenger:
 \begin{eqnarray}
  b^t & = & \argmin{b} \left[\max_{a\neq b} U_1^t(s_1,a) - L_1^t(s_1,b)\right], \label{def:best}\\
  c^t & = & \argmax{c \neq b^t} U_1^t(s_1,c)\label{def:challenger}\,.
 \end{eqnarray}
Then for all remaining steps we follow an optimistic policy, for all $h \in [2,H]$, \[\pi_h^{t+1}(s_h) = \argmax{a} U_h^t(s_h,a).\]
The stopping rule of \OurAlgorithm{} is \begin{equation}\tau = \inf \{ t \in \N : U_1^t(s_1,c^t) - L_1^t(s_1,b^t) \leq \epsilon\}, \label{def:Stopping}\end{equation} and the guess output when stopping is $\guess = b^\tau$. A generic implementation of \OurAlgorithm{} is given in Algorithm~\ref{alg:OurAlgorithm} in Appendix~\ref{app:detail_algo}, where we also discuss some implementation details. Note that, in sharp contrast with the deterministic stopping rule proposed for Sparse Sampling in Lemma~\ref{lemma:SS}, \OurAlgorithm uses an adaptive stopping rule.

\documentclass[../neurips2020.tex]{subfiles}

\section{Analysis of \OurAlgorithm}\label{sec:analysis}

Recall that \OurAlgorithm{} uses policy $\pi^{t+1} = (\pi_1^{t+1}, \dots, \pi_H^{t+1})$ to select the $(t+1)$-st trajectory, $s_1,a_1^{t+1},s_2^{t+1},a_2^{t+1},\dots,s_H^{t+1},a_H^{t+1}$, satisfying $a_h^{t+1} = \pi_h^{t+1}(s_h^{t+1})$ and $s_{h+1}^{t+1} \sim p_h\left(\cdot \left|s_h^{t+1}, a_h^{t+1}\right.\right)$.


\paragraph{High probability event} To define an event $\cE$ that holds with high probability, let $\cE^r$ (resp. $\cE^p$) be the event that the confidence regions for the mean rewards (resp. transition kernels) are correct:
\begin{align*}
  \cE^r &\eqdef \left\{\forall t\in\N^*, \forall h\in[H], \forall (s_h,a_h)\in\cS\times\cA:\ r_h(s_h,a_h) \in \big[\ell_h^t(s_h,a_h) , u_h^t(s_h,a_h)\big] \right\},\\
  \cE^p &\eqdef \left\{\forall t\in\N^*, \forall h\in[H], \forall (s_h,a_h)\in\cS\times\cA:\ p_h(\cdot|s_h,a_h) \in \cC_h^t(s_h,a_h) \right\}\,.
\end{align*}
For a state-action pair $(s_h,a_h)$, let $p_h^{\pi}(s_h,a_h)$ be the probability of reaching it at step $h$ under policy $\pi$, and let $p_h^t(s_h,a_h) = p_h^{\pi^t}(s_h, a_h)$.
We define the \emph{pseudo-counts} of the number of visits of $(s_h,a_h)$ as
$
\bn_h^t(s_h,a_h) \eqdef \sum_{s=1}^t p_h^{s}(s_h,a_h)\,.
$
As $n_h^t(s_h,a_h) - \bn_h^t(s_h,a_h)$ is a martingale, the counts should not be too far from the pseudo-counts. Given a rate function $\beta^{\text{cnt}}$, we define the event
\[
\cE^{\text{cnt}} \eqdef \left\{ \forall t \in \N^\star, \forall h\in [H],\forall (s_h,a_h)\in\cS\times\cA:\ n_h^t(s_h,a_h) \geq \frac{1}{2}\bn_h^t(s_h,a_h)-\beta^{\text{cnt}}(\delta)  \right\}\,.
\]
Finally, we define $\cE$ to be the intersection of these three events: $
\cE = \cE^r\cap \cE^p \cap \cE^{\text{cnt}}$.

\subsection{Correctness} \label{subsec:correctness}
One can easily prove by induction (see Appendix~\ref{app:events}) that
\[\cE^r \cap \cE^p \subseteq \bigcap_{t\in \N^\star} \bigcap_{h=1}^H \Big[ \bigcap_{s_h,a_h} \Big(Q_h(s_h,a_h) \in \left[L_h^t(s_h,a_h) , U_h^t(s_h,a_h)\right]\Big).\]
As the arm $\hat{a}$ output by \OurAlgorithm{} satisfies $L_1(s_1,\hat{a}) > \max_{c \neq \hat{a}} U_1(s_1,c) - \epsilon$, on the event $\cE \subseteq \cE^r \cap \cE^p$ it holds that $Q_1(s_1,\hat{a}) > \max_{c \neq \hat{a}} Q_1(s_1,c) - \epsilon$. Thus \OurAlgorithm{} can only output an $\epsilon$-optimal action. Hence a sufficient condition  for \OurAlgorithm{} to be $(\epsilon,\delta)$-correct is $\bP(\cE) \geq 1 - \delta$.

In Lemma~\ref{lemma:concentration_master_event} below, we provide a calibration of the thresholds functions $\beta^r,\beta^p$ and $\beta^{\text{cnt}}$ such that this sufficient condition holds. This result, proved in Appendix~\ref{app:concentration_events}, relies on new time-uniform concentration inequalities that follow from the method of mixtures \citep{de2004self}.


\begin{lemma}
\label{lemma:concentration_master_event}
For all $\delta\in[0,1]$, it holds that $\PP(\cE) \geq 1-\delta$ for the choices
\begin{align*}
  \beta^r(n,\delta) &= \log(3(\support K)^H/\delta) + \log\big(e(1+n)\big)\,,\ \ \ \beta^{\emph{cnt}}(\delta) =  \log\big(3(\support K)^H/\delta\big)\,,\\
  \text{and } \ \ \beta^p(n,\delta) &= \log\big(3(\support K)^H/\delta\big) + (\support-1)\log\big(e(1+n/(\support-1))\big)\,.
\end{align*}
Moreover, the maximum of these three thresholds defined (by continuity when $B=1$) as
\begin{eqnarray*}
 \beta(n,\delta) & \eqdef & \!\!\!\max_{c \in \{r,p,{\emph{cnt}}\}}\beta^c(n,\delta) =  \log\big(3(BK)^H/\delta\big) + (B-1)\log\big(e(1+n/(B-1))\big) \label{def:master_threshold},
\end{eqnarray*}
is such that $n \mapsto \beta(n,\delta)$ is non-decreasing and $n \mapsto \beta(n,\delta)/n$ is non-increasing.
\end{lemma}

\subsection{Sample Complexity}

In order to state our results, we define the following sub-optimality gaps.
$\Delta_h(s_h,a_h)$ measures the gap in future discounted reward between the optimal action $\pi_h^\star(s_h)$ and the action $a_h$, whereas $\Delta_1^\star(s_1,a_1)$ also takes into account the gap of the second best action and the tolerance level $\epsilon$.

\begin{definition}
\label{def:gaps} Recall that $\Delta = \min_{a\neq a^\star} \left[Q_1(s_1,a^\star) - Q_1(s_1,a)\right]$. For all $h \in [H]$, we let
\begin{eqnarray*}
 \Delta_h(s_h,a_h)& =& Q_h(s_h,\pi_h^\star(s_h)) - Q_h(s_h,a_h),\\
\Delta_1^\star(s_1,a_1) &= &\max\left(\Delta_1(s_1,a_1) ; \Delta ;  \epsilon \right),
\end{eqnarray*}
and we denote $\tilde{\Delta}_h(s_h,a_h) = \left\{\begin{array}{cl} \Delta_1^\star(s_h,a_h), & \text{if } h=1,\\ \Delta_h(s_h,a_h), & \text{if } h\geq 2. \end{array}\right.$
\end{definition}

Our sample complexity bounds follow from the following crucial theorem, which we prove in Appendix~\ref{app:FirstAnalysis}, that relates the pseudo-counts of state-action pairs at time $\tau$ to the corresponding gap.

\begin{theorem} \label{thm:newthm}
If $\cE$ holds, every $(s_h,a_h)$ is such that {\small
\[\bar n_h^{\tau}(s_h,a_h) \tilde{\Delta}_{h}(s_h,a_h) \leq 64\sqrt{2}(1+\sqrt{2}) \left(\sqrt{BK}\right)^{H-h} \sqrt{\bar n^\tau_{h}(s_h,a_h)\beta(\overline{n}_{h}^{\tau}(s_h,a_h), \delta)}.\]}
\end{theorem}

Introducing the constant $C_0 = (64\sqrt{2}(1+\sqrt{2}))^2$ and letting $c_\delta = \log\left(\tfrac{3(BK)^H}{\delta}\right)$,  Lemma~\ref{lemma:technical} stated in Appendix~\ref{proof:technical} permits to prove that, on the event $\cE$, any $(s_h,a_h)$ for which $\tilde{\Delta}_{h}(s_h,a_h) > 0$ satisfies
{\footnotesize\begin{equation}\bar n_h^{\tau}(s_h,a_h) \leq \frac{C_0(BK)^{H\!-  \!h}\!\!}{\tilde{\Delta}^2_{h}(s_h,a_h)}\!\left[c_\delta \! + 2(B\!-\!1)\log\left(\frac{C_0(BK)^{H\!-  \!h}\!\!}{\tilde{\Delta}^2_{h}(s_h,a_h)}\left[\frac{c_\delta}{\sqrt{B\!-\!1}} + 2\sqrt{e(B\!-\!1)}\right]\right)\! +(B\!-\!1)\right]\label{ub:pseudocount}\end{equation}}
As $\tilde{\Delta}_1(s_1,a_1) = \max\left(\Delta_1(s_1,a_1) ; \Delta ;  \epsilon \right)$ is positive, the following corollary follows from summing the inequality over $a_1$, as $\bar n_1^{\tau}(s_1,a_1) = n_1^{\tau}(s_1,a_1)$ and $\tau = \sum_{a_1} n^{\tau}_1(s_1,a_1)$.

\begin{corollary}\label{cor:firstbound}The number of episodes used by MDP-GapE satisfies
\[\bP\left( \tau = \cO\left(\sum_{a_1} \frac{(BK)^{H-1}}{\left(\Delta_1(s_1,a_1) \vee \Delta \vee  \epsilon \right)^2}\left[ \ln\left(\frac{1}{\delta}\right) + BH\ln(BK)\right]\right)\right) \geq 1 -\delta\;.\]
\end{corollary}

The upper bound on the sample complexity $n = H \tau$ of \OurAlgorithm{} that follows from Corollary~\ref{cor:firstbound} improves over the $\mathcal{O}(H^5(BK)^H/\varepsilon^2)$ sample complexity of \SparseSampling{}. It is also smaller than the $\mathcal{O}(H^4(BK)^H/\Delta^2)$ samples needed for \BRUE to have a reasonable upper bound on its simple regret. The improvement is twofold: first, this new bound features the problem dependent gap $\Delta(s_1,a_1) \vee \Delta \vee \varepsilon$ for each action $a_1$ in state $s_1$, whereas previous bounds were only expressed with $\varepsilon$ or $\Delta$. Second, it features an improved scaling in $H^2$. 

It is also possible to provide bounds that features the gaps $\tilde{\Delta}_h(s_h,a_h)$ in the \emph{whole} tree, beyond depth one. To do so, we shall consider \emph{trajectories} $t_{1:H} = (s_1,a_1,\dots,s_H,a_H)$ or trajectory \emph{prefixes} $t_{1:h} = (s_1,a_1,\dots,s_h,a_h)$ for $h \in [H]$. Introducing the probability $p_h^{\pi}(t_{1:h})$ that the prefix $t_{1:h}$ is visited under policy $\pi$, we can further define the pseudo-counts $\bar n_h^{t}(t_{1:h}) = \sum_{s=1}^{t}p_h^{\pi^{s}}(t_{1:h})$. One can easily show that for all $h \in [H]$,
$\bar n_H^{\tau}(t_{1:H}) \leq \bar n_h^{\tau}(t_{1:h}) \leq \bar n_h^{\tau}(s_h,a_h),$
if $(s_h,a_h)$ is the state-action pair visited in step $h$ in the trajectory $t_{1:H}$, and  \eqref{ub:pseudocount} leads to the following upper bound.

\begin{corollary}
\label{cor:trajectorybound}On the event $\cE$, $\bar n_h^{\tau}(t_{1:h}) = \mathcal{O}\left(\left[\min_{\ell=1}^{h} \tfrac{(BK)^{H-\ell}}{\left(\tilde{\Delta}_{\ell}(s_{\ell},a_{\ell})\right)^2}\right]\ln\left(\frac{3(BK)^H}{\delta}\right)\right)$.
\end{corollary}

In particular, using that $\tau = \sum_{t_{1:H} \in \cT} \bar n_h^{\tau}(t_{1:H})$ where $\cT$ is the set of $(BK)^{H}$ complete trajectories leads to a sample complexity bound featuring all gaps. However, its improvement over the bound of Corollary~\ref{cor:firstbound} is not obvious in the general case. For $B=1$, that is for planning in a deterministic MDP with possibly random rewards, a slightly different proof technique leads to the following improved gap-dependent sample complexity bound (see the proof in Appendix~\ref{app:deterministic}).

\begin{theorem}[deterministic case] \label{thm:deterministic} When $B=1$, \OurAlgorithm{} satisfies
\[\bP\left( \tau = \mathcal{O}\left( \sum_{t_{1:H} \in \cT} \left[\min_{h=1}^{H} \frac{\left(\sum_{\ell = h}^{H}\gamma^{\ell}\right)^2}{\left(\tilde{\Delta}_h^2(s_h,a_h)\right)^2}\right]\left( \ln\left(\frac{1}{\delta}\right) + H\ln(K)\right) \right) \right)\geq 1 - \delta.\]
\end{theorem}

\paragraph{Scaling in $\varepsilon$} A majority of prior work on planning in MDPs has obtained sample complexity bounds that scale with $\varepsilon$ only, in the discounted setting. Neglecting the gaps, Corollary~\ref{cor:firstbound} gives a $\mathcal{O}(H^2(BK)^H/\varepsilon^2)$ upper bound that yields a crude $\tilde{\mathcal{O}}\left(\varepsilon^{-\left[2 +\log(BK)/\ln(1/\gamma)\right]}\right)$ sample complexity in the discounted setting in which $H \sim \ln(1/\varepsilon)/\ln(1/\gamma)$. This exponent is larger than that in previous work, which features some notion of near-optimality dimension $\kappa$ (see Table~\ref{tab:different_settings_in_literature}). However, our analysis was not tailored to optimizing this exponent, and we show in Section~\ref{sec:experiments} that the empirical scaling of \OurAlgorithm{} in $\varepsilon$ can be much smaller than the one prescribed by the above crude bound.


\documentclass[../neurips2020.tex]{subfiles}

\section{Numerical Experiments}\label{sec:experiments}

We consider random discounted MDPs with infinite horizon in which the maximal number $B$ of successor states and the sparsity of rewards are controlled. The transition kernel is generated as follows: for each transition in $\mathcal{S} \times \mathcal{A}$, we uniformly pick $B$ next states in $\mathcal{S}$. The cumulative transition probabilities to these states are computed by sorting $B-1$ numbers uniformly sampled in $(0,1)$. The reward kernel is computed by selecting a proportion of the transitions to have non-zero rewards with means sampled uniformly in $(0,1)$. The values for these parameters are shown in Table~\ref{tab:env}.

\begin{table}[htbp]
	\caption{Experimental setting.\label{tab:configuration}}
	\begin{center}
		\begin{subtable}[b]{.35\textwidth}
			\caption{Environment parameters\label{tab:env}}

			\begin{tabular}{lc}
				\toprule
				States $\mathcal{S}$ &  $200$ \\
				Actions $\mathcal{A}$ & $5$  \\
				Number $B$ of successors & $2$ \\
				Reward sparsity & $0.5$ \\
				\bottomrule
			\end{tabular}
		\end{subtable}
		\hspace*{0.2cm}
		\begin{subtable}[b]{.55\textwidth}
			\caption{\OurAlgorithm{} parameters\label{tab:agent}}

			\begin{tabular}{lc}
				\toprule
				Discount factor $\gamma$ &  0.7 \\
				Confidence level $\delta$ & 0.1  \\
				Exploration function $\beta_r(n_h^t, \delta)$ & $\log\frac{1}{\delta} + \log\log n_h^t$ \\
				Exploration function $\beta_p(n_h^t, \delta)$ & $\log\frac{1}{\delta} + \log n_h^t$ \\
				\bottomrule
			\end{tabular}
		\end{subtable}
	\end{center}
\end{table}

\paragraph{Fixed-confidence: Correction and sample complexity}

We verify empirically that \OurAlgorithm{} is ($\varepsilon$, $\delta$)-correct  while stopping with a reasonable number of oracle calls. Table~\ref{tab:agent} shows the choice of parameters for the algorithm. For various values of the desired accuracy $\varepsilon$ and of the corresponding planning horizon $H = \lceil\log_{\gamma}(\varepsilon(1-\gamma)/2)\rceil$ (see Section~\ref{sec:setup}), we run simulations on 200 random MDPs. We report in Table~\ref{tab:pac} the distribution of the number $n = \tau H$ of oracle calls and the simple regret $\bar{r}_n(\hat{a}_n)$ of \OurAlgorithm{} over these 200 runs.  We first observe that \OurAlgorithm{} verifies $\bar{r}_n(\hat{a}_n) < \varepsilon$ in all simulations, despite the use of smaller exploration functions compared to those prescribed in Lemma~\ref{lemma:concentration_master_event}. We then compare its sample complexity to that of \SparseSampling{}, which is deterministic and for which $n_{\text{SS}}$ given in Lemma~\ref{lemma:SS} is a tight upper bond. We see that the sample complexity of \OurAlgorithm{} is an order of magnitude smaller than that of \SparseSampling{}.

\begin{table}[htbp]
	\caption{Simple regret and number of oracle calls, averaged over $200$ simulations}
	\label{tab:pac}
	\centering
	\begin{tabular}{lccccc}
		\toprule
		\multicolumn{1}{l}{\multirow{3}{*}{$\varepsilon$}} &
		\multicolumn{1}{l}{\multirow{3}{*}{$H$}} &
		\multicolumn{3}{c}{\OurAlgorithm}    &
		\SparseSampling \\
		\cmidrule(lr){3-5}
		\cmidrule(lr){6-6}
		&
		&
		\multicolumn{1}{c}{max $r_n$} &
		\multicolumn{1}{c}{median $n$} &
		\multicolumn{1}{c}{max $n$} &
		$n_{\text{SS}}$ \\
		\midrule
		$1$ & $6$ & \num{6e-2} & \num{6.3e3} & \num{1.9e4} &\num{8e9} \\
		$0.5$ & $8$ & \num{4e-3} & \num{5.5e4} & \num{2.2e5} & \num{1e13} \\
		$0.2$ & $10$ & \num{2.9e-3} & \num{3.4e5} & \num{2.3e6} & \num{3e16} \\
		\bottomrule
	\end{tabular}
\end{table}


\paragraph{Scaling in $\varepsilon$}

As discussed above, Corollary~\ref{cor:firstbound} with the aforementioned choice of the planning horizon, yields a crude sample complexity bound on the
$\tilde{\mathcal{O}}\left(\varepsilon^{-\left[2 +\log(BK)/\ln(1/\gamma)\right]}\right) = \tilde{\mathcal{O}}\left((1/\varepsilon)^{8.4}\right)$ in our experimental setting.
However, we observe that the empirical exponent can be much smaller in practice: plotting the average sample complexity $n$ (estimated over the 200 MDPs) as a function of $\log(1/\varepsilon)$ in Figure~\ref{fig:effective-complexity} and measuring the slope of the curve yields $n \simeq \mathcal{O}\big(\left(1/\varepsilon\right)^{3.9}\big)$.

\paragraph{Comparison to the state of the art}

In the fixed-confidence setting, most existing algorithms are considered theoretical and cannot be applied to practical cases. For instance, for our problem with $K=5$ and $\varepsilon=1$, \SparseSampling~\citep{Kearns02SS} and \texttt{SmoothCruiser}~\citep{SmoothCruiser19} both require a fixed budget\footnote{In non-regularized MDPs, \texttt{SmoothCruiser} has the same sample complexity as \SparseSampling.} of at least $n_{\text{SS}}=\num{8e9}$. Likewise, \texttt{Trailblazer} \citep{TrailBlazer16} is a recursive algorithm which did not terminate in our setting. We did not implement \STOP \cite{STOP14} as it requires to store a tree of policies, which is very costly even for moderate horizons. In comparison, Table~\ref{tab:pac} shows that \OurAlgorithm{} stopped after $n=\num{1.9e4}$ oracle calls in the worst case. To the best of our knowledge, \OurAlgorithm{} is the first $(\varepsilon,\delta)$-correct algorithm for general MDPs with an easy implementation and a reasonable running time in practice.
The only planning algorithms that can be run in practice are in the fixed-budget setting, which we now consider.

\paragraph{Fixed-budget evaluation}

We compare \OurAlgorithm{} to three existing baselines: first, the \KLOLOP{} algorithm \citep{leurent2019practical}, which uses the same upper-confidence bounds on the rewards $u_h^t$ and states values $U_h^t$ as \OurAlgorithm{}, but is restricted to \emph{open-loop} policies, i.e. sequences of actions only. Second, the \BRUE{} algorithm \citep{Feldman14BRUE} which explores uniformly and handles closed-loop policies. Third, the popular \UCT{} algorithm \citep{Kocsis06UCT}, which is also closed-loop and performs optimistic exploration at all depths. \UCT and its variants lack theoretical guarantees, but they have been shown successful empirically in many applications. For each algorithm, we tune the planning horizon $H$ similarly to \KLOLOP{}, by dividing the available budget $n$ into $\tau$ episodes, where $\tau$ is the largest integer such that $\tau \log \tau/(2\log 1/\gamma)\leq n$, and choose $H = \log \tau/(2\log 1/\gamma)$. The exploration functions are those of \KLOLOP{} and depend on $\tau$: $\beta_r(n_h^t,\delta) = \beta_p(n_h^t, \delta) = \log(\tau)$. Again, we perform 200 simulations and report in Figure~\ref{fig:fixed-budget} the mean simple regret, along with its $95\%$ confidence interval. We observe that \OurAlgorithm{} compares favourably with these baselines in the high-budget regime.

\begin{figure}[htbp]
	\centering
	\begin{minipage}{0.49\textwidth}
		\centering
	    \includegraphics[width=0.95\textwidth]{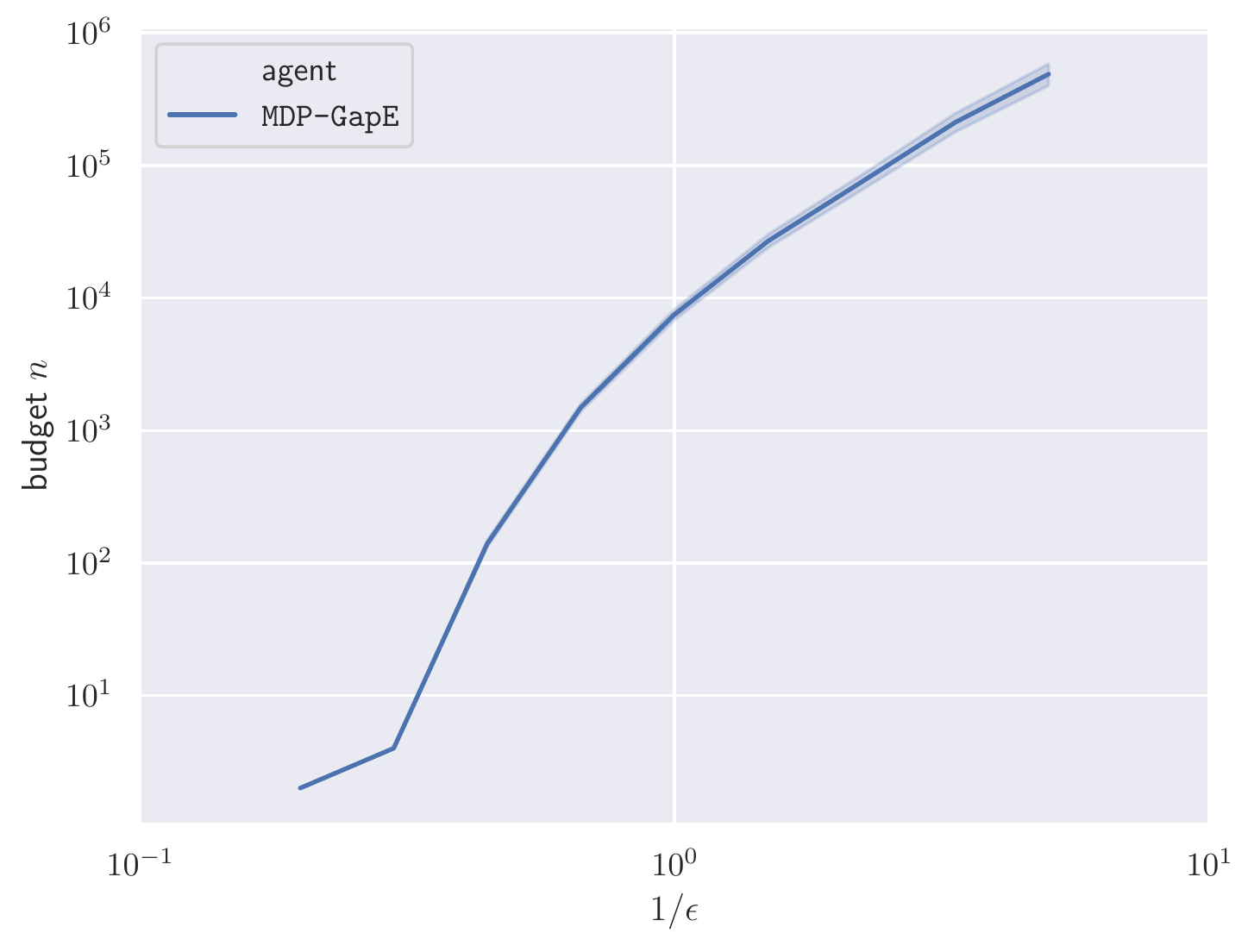}
	    \caption{Polynomial dependency of the number $n$ of oracle calls with respect to $1/\varepsilon$.}
		\label{fig:effective-complexity}
	\end{minipage}\hfill
	\begin{minipage}{0.49\textwidth}
		\centering
		\includegraphics[width=0.95\textwidth]{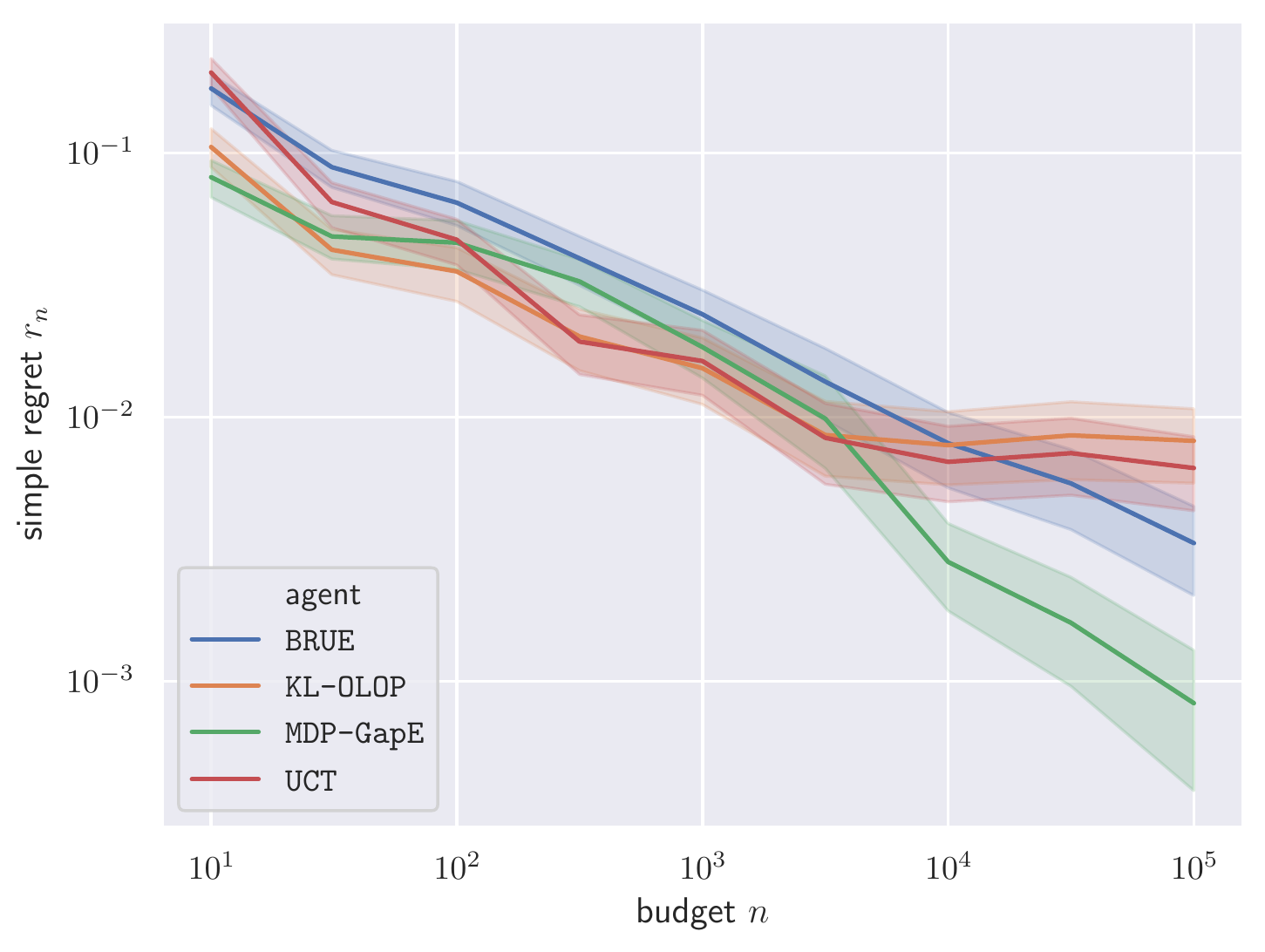}
		\caption{Comparison to \KLOLOP in a fixed-budget setting.}
		\label{fig:fixed-budget}
	\end{minipage}

\end{figure}

\section{Conclusion}

We proposed a new, efficient algorithm for Monte-Carlo planning in Markov Decision Processes, that combines tools from best arm identification and optimistic planning and exploits tight confidence regions on mean rewards and transitions probabilities. We proved that \OurAlgorithm{} attains the smallest existing gap-dependent sample complexity bound for general MDPs with stochastic rewards and transitions, when the branching factor $B$ is finite.
In future work, we will investigate the worse-case complexity of \OurAlgorithm{}, that is try to derive an upper bound on its sample complexity that only
features $\varepsilon$ and some appropriate notion of near-optimality dimension.

\section*{Acknowledgments}

Anders Jonsson is partially supported by the Spanish grants TIN2015-67959 and PCIN-2017-082.

%



\bibliography{neurips2020_bib}

\newpage
\appendix
\newpage
\section{Detailed Algorithm}
\label{app:detail_algo}
In this section we provide a detailed algorithm for \OurAlgorithm, namely Algorithm~\ref{alg:OurAlgorithm}.

\begin{algorithm}[h!]
  \caption{\OurAlgorithm}
  \label{alg:OurAlgorithm}
    \begin{algorithmic}[1]
      \STATE {\bfseries Input:} confidence level $\delta$, tolerance $\epsilon$
      \STATE initialize data lists $\cD_h \gets [ \ ]$ for all $h \in [H]$
      \FOR{$t=1 \ldots$}
      \STATE //Update confidence bounds
      \STATE $U_h^{t-1},L_h^{t-1} \gets \UpdateBounds(t,\delta,\cD_h)$
      \IF{$U_1^{t-1}(s_1,c^t) - L_1^{t-1}(s_1,b^t) \leq \epsilon$}
        \STATE{\bfseries return} $b_{t-1}$, {\bfseries break}
      \ENDIF
      \STATE // Best
      \STATE $b^{t-1}  \gets  \argmin{b} \left[\max_{a\neq b} U_1^{t-1}(s_1,a) - L_1^{t-1}(s_1,b)\right]$
      \STATE //Challenger
      \STATE $c^{t-1}  \gets  \argmax{c \neq b^t} U_1^{t-1}(s_1,c)$
      \STATE //Exploration
      \STATE $a_1^t \gets \argmax{a \in \{b^{t-1},c^{t-1}\}} \left[U_1^{t-1}(s_1,a) - L_1^{t-1}(s_1,a)\right]$
      \STATE observe reward $r_{1}^t$, next state $s_{2}^t$, save $\data_1.\mathrm{append}(s_{1}^t, a_{1}^t, s_{2}^t, r_{1}^t)$
      \FOR{step $h=2, \ldots, H$}
        \STATE $a_h^t \gets \argmax{a} U_h^{t-1} (s_h^t, a)$
        \STATE observe reward $r_{h-1}^t$, next state $s_{h}^t$,save $\data_h.\mathrm{append}(s_{h}^t, a_{h}^t, s_{h+1}^t, r_{h}^t)$
      \ENDFOR
      \ENDFOR
    \end{algorithmic}
\end{algorithm}

\paragraph{Implementation details}
 There are different ways to store and update the confidence bounds on the $Q$-value (that is, to specify the \texttt {UpdateBounds} subroutine) according to how we merge information across states. 
 
 The most obvious one, suggested by previous work \citep{bubeck2010open,leurent2019practical,busoniu2012optimistic} (and also implemented for our experiments) does not merge information at all and builds a search \emph{tree} in which a node $(s_h,a_h)$ at depth $h$ is identified with the sequence of $h$ states and actions that leads to it. It leads to a very simple update: after each trajectory, one only needs to update the confidence bounds, $U_h(s_h,a_h)$ and $L_h(s_h,a_h)$, of the visited action-state pairs.
Another option is to merge information for the same states and a fixed depth. But in this case the search tree becomes a \emph{graph} and after each trajectory we need to re-compute the values $U_h(s_h,a_h)$ for all stored state action pairs $(s_h,a_h)$ at each depth.

\section{Correctness of \OurAlgorithm}
\label{app:events}

In this section we prove the correctness of \OurAlgorithm under the assumption that the event $\cE^r \cap \cE^p$ holds. Concretely, we prove by induction that
\[\cE^r \cap \cE^p \subseteq \bigcap_{t\in \N^\star} \bigcap_{h=1}^H \Big[ \bigcap_{s_h,a_h} \Big(Q_h(s_h,a_h) \in \left[L_h^t(s_h,a_h) , U_h^t(s_h,a_h)\right]\Big)\Big].\]
The base case is given by $h=H+1$, in which case by our previous convention,
\begin{align*}
L_{H+1}^t(\cdot,\cdot) = Q_{H+1}(\cdot,\cdot) = U_{H+1}^t(\cdot,\cdot) = 0.
\end{align*}
For the inductive case, assume that the inclusion holds at depth $h+1$. Then we have
\begin{align*}
L_h^t(s_h,a_h) &= \ell_h^t(s_h,a_h) + \gamma \min_{p \in \cC_h^t(s_h,a_h)} \sum_{s'} p(s'|s_h,a_h) \max_{a'} L_{h+1}^t(s',a')\\
 &\leq \ell_h^t(s_h,a_h) + \gamma \sum_{s'} p_h(s'|s_h,a_h) \max_{a'} L_{h+1}^t(s',a')\\
 &\leq r_h(s_h,a_h) + \gamma \sum_{s'} p_h(s'|s_h,a_h) Q_{h+1}(s',\arg\max_{a'} L_{h+1}^t(s',a'))\\
 &\leq r_h(s_h,a_h) + \gamma \sum_{s'} p_h(s'|s_h,a_h) Q_{h+1}(s',\pi_{h+1}^*(s')) = Q_h(s_h,a_h)\\
 &\leq u_h^t(s_h,a_h) + \gamma \sum_{s'} p_h(s'|s_h,a_h) \max_{a'} U_{h+1}^t(s',a')\\
 &\leq u_h^t(s_h,a_h) + \gamma \max_{p \in \cC_h^t(s_h,a_h)} \sum_{s'} p(s'|s_h,a_h) \max_{a'} U_{h+1}^t(s',a') = U_h^t(s_h,a_h),
\end{align*}
where we have used $r_h(s_h,a_h) \in \big[\ell_h^t(s_h,a_h) , u_h^t(s_h,a_h)\big]$ and $p_h(\cdot|s_h,a_h) \in \cC_h^t(s_h,a_h)$.

\section{Concentration Events}
\label{app:concentration_events}
In this section we prove that the event $\cE$ holds with high probability. But before we need several concentration inequalities.
\subsection{Deviation Inequality for Categorical Distributions}
Let $X_1, X_2,\ldots,X_n,\ldots$ be i.i.d. samples from a distribution supported over $\{1,\ldots,m\}$, of probabilities given by $p\in\Sigma_m$, where $\Sigma_m$ is the probability simplex of dimension $m-1$. We denote by $\hp_n$ the empirical vector of probabilities, i.e. for all $k\in\{1,\ldots,m\}$
\[
\hp_{n,k} = \frac{1}{n} \sum_{\ell=1}^n \ind(X_\ell = k)\,.
\]
Note that  an element $p \in \Sigma_m$ will sometimes be seen as an element of $\R^{m-1}$ since $p_m = 1- \sum_{k=1}^{m-1} p_k$. This should be clear from the context. We denote by $H(p)$ the (Shannon) entropy of $p\in\Sigma_m$,
\[
H(p) = \sum_{k=1}^m p_k \log(1/p_k)\,.
\]
\begin{proposition}
For all $p\in\Sigma_m$, for all $\delta\in[0,1]$,
\begin{equation*}
    \PP\Big(\exists n\in \N^*,\, n\KL(\hp_n, p)> \log(1/\delta) + (m-1)\log\big(e(1+n/(m-1))\big)\Big)\leq \delta\,.
\end{equation*}
\label{prop:max_ineq_categorical}
\end{proposition}

\begin{proof}
We apply the method of mixture with a Dirichlet prior on the mean parameter of the exponential family formed by the set of categorical distribution on  $\{1,\ldots,m\}$. Letting
\[\phi_p(\lambda) = \log \bE_{X\sim p}\left[e^{\lambda X}\right] =\log(p_m+\sum_{k=1}^{m-1}p_k e^{\lambda_k}),\]
be the log-partition function, the following quantity is a martingale:
\[
M_n^\lambda = e^{n \langle \lambda,\hp_n\rangle - n \phi_p(\lambda)}.
\]
We set a Dirichlet prior $q\sim \Dir(\alpha)$ with $\alpha \in {\R^*_+}^m$ and for $\lambda_q = (\nabla\phi_p)^{-1}(q)$ and consider the integrated martingale
\begin{align*}
    M_n &= \int M_n^{\lambda_q} \frac{\Gamma\Big(\sum_{k=1}^m \alpha_k\Big)}{\prod_{k=1}^m\Gamma(\alpha_k)} q_k^{\alpha_k-1}\diff q\\
    &= \int e^{n\big(\KL(\hp_n,p)-\KL(\hp_n,q)\big)}\frac{\Gamma\Big(\sum_{k=1}^m \alpha_k\Big)}{\prod_{k=1}^m\Gamma(\alpha_k)} q_k^{\alpha_k-1}\diff q\\
    &=  e^{n\KL(\hp_n,p) + n H(\hp_n)} \int \frac{\Gamma\Big(\sum_{k=1}^m \alpha_k\Big)}{\prod_{k=1}^m\Gamma(\alpha_k)} q_k^{n\hp_{n,k}+\alpha_k-1}\diff q\\
    &= e^{n\KL(\hp_n,p) + n H(\hp_n)}\frac{\Gamma\Big(\sum_{k=1}^m \alpha_k\Big)}{\prod_{k=1}^m\Gamma(\alpha_k)} \frac{\prod_{k=1}^m\Gamma(\alpha_k + n \hp_{n,k})}{\Gamma\Big(\sum_{k=1}^m \alpha_k + n\Big)}\,,
\end{align*}
where in the second inequality we used Lemma~\ref{lem:martingale_to_diff_kl}. Now we choose the uniform prior $\alpha= (1,\ldots,1)$. Hence we get
\begin{align*}
M_n &= e^{n\KL(\hp_n,p) + n H(\hp_n)} (m-1)! \frac{\prod_{k=1}^m\Gamma(1 + n \hp_{n,k})}{\Gamma( m + n)}\\
&= e^{n\KL(\hp_n,p) + n H(\hp_n)} (m-1)! \frac{\prod_{k=1}^m(n \hp_{n,k})!}{n!} \frac{n!}{(m+n-1)!}\\
& = e^{n\KL(\hp_n,p) + n H(\hp_n)} \frac{1}{\binom{n}{n \hp_n}} \frac{1}{\binom{m+n-1}{m-1}}.
\end{align*}
Thanks to Theorem 11.1.3 by \citep{cover2012elements} we can upper bound the multinomial coefficient as follows: for $M\in \N^*$ and $x\in \{0,\ldots,M\}^m$ such that $\sum_{k=1}^m x_k = M$ it holds
\[
\binom{M}{x} = \frac{M!}{\prod_{k=1}^m x_k!} \leq e^{M H(x/M)}\,.
\]
Using this inequality we obtain
\begin{eqnarray*}
M_n &\geq&  e^{n\kl(\hp_n,p) +n H(\hp_n) - n H(\hp_n) -(m+n-1) H\big((m-1)/(m+n-1)\big)} \\
&=&  e^{n\KL(\hp_n,p)-(m+n-1) H\big((m-1)/(m+n-1)\big)}\,.
\end{eqnarray*}
It remains to upper-bound the entropic term
\begin{align*}
   (m+n-1) H\big((m-1)/(m+n-1)\big) &= (m-1) \log \frac{m+n-1}{m-1} + n \log \frac{m+n-1}{n}\\
   &\leq  (m-1) \log\big( 1 + n/(m-1)\big) + n \log(1+(m-1)/n)\\
   &\leq (m-1) \log\big( 1 + n/(m-1)\big) + (m-1)\,.
\end{align*}
Thus we can lower bound the martingale as follows
\begin{align*}
    M_n &\geq e^{n\KL(\hp_n,p)}\big( e(1+n/(m-1))  \big)^{m-1}\,.
\end{align*}
Using the fact that, for any supermartingale it holds that
\begin{equation}\bP\left(\exists n \in \N^* : M_n > 1/\delta\right) \leq \delta \bE[M_1],\label{eq:supermartingale}\end{equation}
which is a well-known property used in the method of mixtures (see \citep{de2004self}), we conclude that
\[
\PP\Big(\exists n\in \N^*,\, n\KL(\hp_n, p)> (m-1)\log\big(e(1+n/(m-1))\big)+ \log(1/\delta)\Big) \leq \delta\,.
\]
\end{proof}

\begin{lemma}
\label{lem:martingale_to_diff_kl}
For $q,p \in \Sigma_m$ and $\lambda\in \R^{m-1}$,
\[
\langle \lambda,q \rangle -\phi_{p}(\lambda) = \KL(q,p) -  \KL(q,p^\lambda)\,,
\]
where $\phi_p(\lambda) = \log(p_m+\sum_{k=1}^{m-1}p_k e^{\lambda_k})$ and $p^\lambda = \nabla \phi_{p_0}(\lambda)$.
\end{lemma}
\begin{proof}
 There is a more general way than the ad hoc one below to prove the result. First note that
 \[
 p^\lambda_k = \frac{p_k e^{\lambda_k}}{p_m +  \sum_{\ell=1}^{m-1} p_\ell e^{\lambda_\ell}}\,,
 \]
 which implies that
 \[
 p_m + \sum_{k=1}^{m-1} p_k e^{\lambda_k} = \frac{p_m}{p_m^\lambda}, \qquad \lambda_k = \log\frac{p_k^\lambda}{p_k} + \log \frac{p_m}{p_m^\lambda}\,.
 \]
 Therefore we get
 \begin{align*}
 \langle \lambda, q\rangle -\phi_p (\lambda) &= \sum_{k=1}^{m-1} q_k \log \left(\frac{p_k^\lambda}{p_k} \frac{p_m}{p_m^\lambda}\right) - \log\left(p_m + \sum_{k=1}^{m-1} p_k e^{\lambda_k}\right)\\
 &= \sum_{k=1}^{m-1} q_k \log\frac{p_k^\lambda}{p_k} +(1-q_m) \log \frac{p_m}{p_m^\lambda} - \log\frac{p_m}{p_m^\lambda}\\
 &= \sum_{k=1}^m q_k \log\frac{p_k^\lambda}{p_k} = \KL(q,p)-\KL(q,p^\lambda)\,.
 \end{align*}
\end{proof}
\subsection{Deviation Inequality for Bounded Distribution}
Let $X_1, X_2,\ldots,X_n,\ldots$ be i.i.d. samples from a distribution $\nu$ of mean $\mu$ supported on $[0,1]$. We denote by $\hmu_n$ the empirical mean
\[
\hmu_{n} = \frac{1}{n} \sum_{\ell=1}^n X_\ell\,.
\]
It is well known, see \citep{garivier2011kl}, that we can "project" the distribution $\nu$ on a Bernoulli distribution with the same mean and then use deviation inequality for Bernoulli to concentrate the empirical mean. This method dos not lead to the sharpest confidence intervals but it provides a good trade-off between complexity computation and accuracy.
\begin{proposition}
  \label{prop:max_ineq_bounded}
  For all distribution $\nu$ of mean $\mu$ supported on the unit interval, for all $\delta\in[0,1]$,
  \[
  \PP\left(\exists n\in\N^*,\, n\kl(\hmu_n,\mu)  > \log(1/\delta) + \log\big(e(1+n)\big)\right) \leq \delta\,.
  \]
\end{proposition}
\begin{proof}
First note that we can upper bound the log-partition function of $\nu$ by the one of a Bernoulli $\Ber(\mu)$, for all $\lambda \in \R$,
\[
\log\left(\EE[e^{\lambda X_n }]\right)\leq \log\left( \EE[X_n e^{\lambda}+ 1-X_n] \right) =\log\left(1-\mu+\mu e^{\lambda} \right)=\phi_\mu(\lambda).
\]
Then we can follow the proof of Proportion~\ref{prop:max_ineq_categorical} with $m=2$ and where $M_n^\lambda$ is only a supermartingale but this does not change the result as the property \eqref{eq:supermartingale} still holds. Thus the proposition follows by specifying Proposition~\ref{prop:max_ineq_categorical} to the case $m=2$.
\end{proof}
\subsection{Deviation Inequality for sequence of Bernoulli Random Variables}
Let $X_1, X_2,\ldots,X_n,\ldots$ be a sequence of Bernoulli random variables adapted to the filtration $(\cF_t)_{t\in\N}$. We restate here Lemma F.4. of \citep{dann2017unifying}.
\begin{proposition}
\label{prop:max_ineq_bernoulli_martingale}
If we denote $p_n= \PP(X_n =1|\cF_{n-1})$, then for all $\delta\in(0,1]$
\[
\PP\left(\exists n \in \N^*: \sum_{\ell=1}^n X_\ell < \sum_{\ell=1}^n p_\ell/2 - \log(1/\delta)\right) \leq \delta \,.
\]
\end{proposition}
\subsection{Proof of Lemma~\ref{lemma:concentration_master_event}}
We just prove that each event forming $\cE = \cE^r\cap \cE^p \cap \cE^n$ holds with high probability. For the first one using Proposition~\ref{prop:max_ineq_bounded}, since the reward are bounded in the unit interval we have
\begin{align*}
\PP\!\big((\cE^r)^c\big) &\leq \sum_{h\in[H]}\!\!\sum_{(s_h,a_h)\in\cS\times\cA}\!\!\! \PP\!\left(\exists t\in\N^*:\  n_h^t(s_h,a_h)\kl\!\big(\hat{r}_h^t(s_h,a_h),r_h(s_h,a_h)\big) > \beta_r(n_h^t(s_h,a_h), \delta) \right)\\
&\leq  \sum_{h\in[H]}\sum_{(s_h,a_h)\in\cS\times\cA} \frac{\delta}{ 3{AS}^H } \leq \frac{\delta}{3}\,.
\end{align*}
where we used Doob's optional skipping in the second inequality in order to apply Proposition~\ref{prop:max_ineq_bounded}, see Section 4.1 of \citep{garivier2018kl}.
Similarly for the confidence regions for the probabilities transitions, using Proposition~\ref{prop:max_ineq_categorical} we obtain
\begin{align*}
\PP\!\big((\cE^p)^c\big) &\leq \sum_{h\in[H]}\!\!\sum_{(s_h,a_h)\in\cS\times\cA}\!\!\! \PP\!\left(\exists t\in\N^*:\  n_h^t(s_h,a_h) \KL\!\big(\hp_h^t(\cdot|s_h,a_h),p_h(\cdot|s_h,a_h)\big) > \beta_p(n_h^t(s_h,a_h), \delta) \right)\\
&\leq  \sum_{h\in[H]}\sum_{(s_h,a_h)\in\cS\times\cA} \frac{\delta}{ 3{AS}^H } \leq \frac{\delta}{3}\,.
\end{align*}
It remains to control the counts, using Proposition~\ref{prop:max_ineq_bernoulli_martingale},
\begin{align*}
\PP\!\big((\cE^{\text{cnt}})^c\big) &\leq \sum_{h\in[H]}\sum_{(s_h,a_h)\in\cS\times\cA} \PP\!\left(\exists t\in\N^*:\  n_h^t(s_h,a_h) <  \frac{1}{2} \bn_h^t(s_h,a_h)-\beta^{\text{cnt}}(\delta)  \right)\\
&\leq  \sum_{h\in[H]}\sum_{(s_h,a_h)\in\cS\times\cA} \frac{\delta}{ 3{AS}^H } \leq \frac{\delta}{3}\,,
\end{align*}
where we used that by definition of the pseudo-counts
\[
\bn_h^t(s_h,a_h) = \sum_{\ell = 1}^t \PP\big((s_h^\ell,a_h^\ell) = (s_h,a_h) |\cF_{\ell-1}\big)\,,
\]
and $\cF_{\ell-1}$ is the information available to the agent at step $\ell$. An union bound allows us to conclude
\[
\PP(\cE^c) \leq \PP\!\big((\cE^r)^c\big)+\PP\!\big((\cE^p)^c\big)+\PP\!\big((\cE^{\text{cnt}})^c\big)\leq \delta\,.
\]

\section{Proof of Theorem~\ref{thm:newthm}}
\label{app:FirstAnalysis}

In this section we present the proof of Theorem~\ref{thm:newthm}, which relies on three important ingredients. The first ingredient is Lemma~\ref{lem:allinone} in Appendix~\ref{app:gaps}, which provides a relationship between the state-action gaps and the diameter $D_h^t(s_h,a_h) \eqdef U_h^t(s_h,a_h) - L_h^t(s_h,a_h)$ of the confidence intervals. The second ingredient is Lemma~\ref{lemma:difference} in Appendix~\ref{app:diameter}, which provides an upper bound on the diameter $D_h^t(s_h,a_h)$. The third ingredient is Lemma~\ref{lemma:technicalCounts} in Appendix~\ref{app:counts}, which relates the actual counts of state-action pairs to the corresponding pseudo-counts. After providing these ingredients, we present the detailed proof of Theorem~\ref{thm:newthm} in Appendix~\ref{subsec:detail}.

\subsection{Relating state-action gaps to diameters}
\label{app:gaps}

Before stating Lemma~\ref{lem:allinone}, we prove an important property of the \UGapE{} algorithm. We recall that $b^t$ and $c^t$ are the candidate best action and its challenger, defined as
\begin{align*}
b^t &= \argmin{b} \left[\max_{a\neq b} U_1^t(s_1,a) - L_1^t(s_1,b)\right],\\
c^t &= \argmax{c \neq b^t} U_1^t(s_1,c)\label{def:challenger}.
\end{align*}
The policy at the root is then defined as $\pi^{t+1}_1(s_1) = \argmax{b \in \{b^t,c^t\}} \left[U_1^t(s_1,b) - L_1^t(s_1,b)\right]$.

\begin{lemma}\label{lem:propUGapE}
For all $t\in[\tau_\delta-1]$, the following inequalities hold:
\begin{enumerate}
 \item $U^t_1\left(s_1,c^t\right) - L^t_1\left(s_1,b^t\right) \leq U^t_1\left(s_1,\pi^{t+1}(s_1)\right) - L^t_1\left(s_1,\pi^{t+1}(s_1)\right)$,
 \item $U^t_1\left(s_1,b^t\right) - L^t_1\left(s_1,c^t\right) < 2\left[U^t_1\left(s_1,\pi^{t+1}(s_1)\right) - L^t_1\left(s_1,\pi^{t+1}(s_1)\right)\right]$.
\end{enumerate}
\end{lemma}

\begin{proof}
We show the first part by contradiction. If the inequality does not hold, we obtain
\begin{align*}
U_1^t(s_1,b^t) - L_1^t(s_1,b^t) &\leq U_1^t(s_1,\pi_1^{t+1}(s_1)) - L_1^t(s_1,\pi_1^{t+1}(s_1)) < U_1^t(s_1,c^t) - L_1^t(s_1,b^t),\\
U_1^t(s_1,c^t) - L_1^t(s_1,c^t) &\leq U_1^t(s_1,\pi_1^{t+1}(s_1)) - L_1^t(s_1,\pi_1^{t+1}(s_1)) < U_1^t(s_1,c^t) - L_1^t(s_1,b^t)\\
 &= \max_{a\neq b^t}U_1^t(s_1,a) - L_1^t(s_1,b^t) \leq \max_{a\neq c^t} U_1^t(s_1,a) - L_1^t(s_1,c^t),
\end{align*}
where the last inequality follows from the definition of $b^t$. Combining the two inequalities yields $U_1^t(s_1,b^t)<U_1^t(s_1,c^t)<\max_{a\neq c^t} U_1^t(s_1,a)$, which contradicts the definition of $c^t$.

For the second part, if $t<\tau_\delta$ then the algorithm has not yet stopped, implying
\begin{align*}
U_1^t(s_1,b^t) - L_1^t(s_1,c^t) &= U_1^t(s_1,b^t)-L_1^t(s_1,b^t)+U_1^t(s_1,c^t)-L_1^t(s_1,c^t)\\
 & \;\;\;\; -\left[U_1^t(s_1,c^t)-L_1^t(s_1,b^t)\right]\\
 &< 2\left[U_1^t(s_1,\pi_1^{t+1}(s_1))-L_1^t(s_1,\pi_1^{t+1}(s_1)\right] - \epsilon.
\end{align*}
\end{proof}

As a consequence of Lemma~\ref{lem:propUGapE}, we can upper bound any confidence interval involving $b^t$ and $c^t$.
\begin{corollary}\label{cor:propUGapE}
For each pair of actions $a,a'\in\{b^t,c^t\}$, it holds that
\[U^t_1\left(s_1,a\right) - L^t_1\left(s_1,a'\right) \leq 2\left[U^t_1\left(s_1,\pi^{t+1}(s_1)\right) - L^t_1\left(s_1,\pi^{t+1}(s_1)\right)\right].\]
\end{corollary}

We are now ready to state Lemma~\ref{lem:allinone}.

\begin{lemma}\label{lem:allinone}
 If $\cE$ holds and $t < \tau_\delta$, for all  $h\in[H]$ and $s_h\in\cS_h(\pi^{t+1})$,
\[\tilde{\Delta}_h(s_h,\pi_h^{t+1}(s_h)) \leq 2\left[U_h^t(s_h,\pi_h^{t+1}(s_h)) - L_h^t(s_h,\pi_h^{t+1}(s_h))\right].\]
\end{lemma}

\begin{proof}
The proof for $h\in[2,H]$ is immediate from the correctness of the confidence bounds implied by $\cE$, and the fact that the selection is optimistic:
\begin{align*}
\Delta_h(s_h,\pi_h^{t+1}(s_h)) &= Q_h(s_h,\pi_h^\star(s_h)) - Q_h(s_h,\pi_h^{t+1}(s_h))\\
 &\leq \max_a U_h^t(s_h,a) - L_h^t(s_h,\pi_h^{t+1}(s_h)) = U_h^t(s_h,\pi_h^{t+1}(s_h)) - L_h^t(s_h,\pi_h^{t+1}(s_h)).
\end{align*}

\noindent
For $h=1$, we prove separately that each term in the max is smaller that the right hand side of desired inequality, that is
\[
\max\left(\Delta_1\big(s_1,\pi^{t+1}(s_1)\big) ; \Delta;  \epsilon \right) \leq 2\left[U_h^t(s_h,\pi_h^{t+1}(s_h)) - L_h^t(s_h,\pi_h^{t+1}(s_h))\right].
\]
Now, by definition of the stopping rule, if $t<\tau_\delta$, $U_1^t\left(s_1,c^t\right) - L_1^t\left(s_1,b^t\right) > \epsilon$. Using the first property in Lemma~\ref{lem:propUGapE} yields
\begin{equation}
\epsilon < U_1^t\left(s_1,\pi^{t+1}(s_1)\right) - L_1^t\left(s_1,\pi^{t+1}(s_1)\right).   \label{eq:Delta1First}
\end{equation}
Then, exploiting the fact that the action with largest UCB is either $b^t$ or $c^t$, it holds on $\cE$ that
\begin{eqnarray*}
 \Delta_1\left(s_1,\pi^{t+1}(s_1)\right) & = & Q_1\left(s_1,a^\star\right) - Q_1\left(s_1,\pi^{t+1}(s_1)\right) \\
 & \leq & \max_{a} U_1^t\left(s_1,a\right) - L_1^t\left(s_1,\pi^{t+1}(s_1)\right) \\
 & = & \max_{a \in \{b^t,c^t\}} U_1^t\left(s_1,a\right) - L_1^t\left(s_1,\pi^{t+1}(s_1)\right).
\end{eqnarray*}
Using Corollary~\ref{cor:propUGapE} to further upper bound the right hand side yields
\begin{equation}
 \Delta_1\left(s_1,\pi^{t+1}(s_1)\right) < 2\left[U_1^t\left(s_1,\pi^{t+1}(s_1)\right) - L_1^t\left(s_1,\pi^{t+1}(s_1)\right)\right]. \label{eq:Delta1Second}
\end{equation}
Finally, one can also write, on the event $\cE$,
\begin{eqnarray*}\Delta = \min_{a\neq a^\star}\left[Q_1(s_1,a^\star) - Q_1(s_1,a)\right] &\leq&  U_1^t(s_1,a^\star) - \max_{a\neq a^\star} \ Q_1(s_1,a) \\
& \leq & \max_{a' \in \{b^t,c^t\}} U_1^t(s_1,a') - \min_{a \in \{b^t,c^t\}} Q_1(s_1,a) \\
& \leq & \max_{a' \in \{b^t,c^t\}} U_1^t(s_1,a') - \min_{a \in \{b^t,c^t\}} L_1^t(s_1,a).
\end{eqnarray*}
In each of the four possible choices of $(a,a')$, Corollary~\ref{cor:propUGapE} implies that 
\begin{equation} \Delta\leq 2 \left[U_1^t\left(s_1,\pi^{t+1}(s_1)\right) - L_1^t\left(s_1,\pi^{t+1}(s_1)\right)\right].\label{eq:DeltaLast}\end{equation}
Lemma~\ref{lem:allinone} follows by combining \eqref{eq:Delta1First}, \eqref{eq:Delta1Second} and \eqref{eq:DeltaLast} with the definition of $\Delta_1^\star\left(s_1,\pi^{t+1}(s_1)\right)$.
\end{proof}

\subsection{Upper bounding the diameters}
\label{app:diameter}

In this section we state and prove Lemma~\ref{lemma:difference}. We use the notation $\sigma_h=\sum_{i=0}^{h-1}\gamma^i$ to upper bound the discounted reward in $h$ steps. As a first step, we prove the following auxiliary lemma. 

\begin{lemma}\label{lemma:KLball}
If $\cE$ holds, for each $h\in[H]$, each $(s_h,a_h)$ and each $q \in \cC_h^t(s_h,a_h)$,
\[
\sum_{s'} \left( q(s'|s_h,a_h) - p_h(s'|s_h,a_h) \right) U_{h+1}^t(s',\pi_{h+1}^{t+1}(s')) \leq 2\sqrt{2}\sigma_{H-h} \sqrt{ \frac{ \beta( n_h^t(s_h,a_h), \delta) } {n_h^t(s_h,a_h) \vee 1} }.
\]
\end{lemma}

\begin{proof}
First note that for each state $s'$, $U_{h+1}^t(s',\pi_{h+1}^{t+1}(s'))$ can be expressed as an expectation on the form $\mathbb{E}^{\pi^{t+1}} \big\{ \sum_{i=h+1}^H \gamma^{i-h-1} u_i^t(s_i,a_i) \mid s_{h+1}=s' \big\}$, which is upper bounded by $\sum_{i=h+1}^H \gamma^{i-h-1} = \sigma_{H-h}$ since $u_i^t(s_i,a_i)\leq 1$ for each $(s_i,a_i)$. Note that for $h=H$, $\sigma_{H-H}=\sigma_0=0$. If $n_h^t(s_h,a_h)=0$ the result trivially holds by the conventions adopted for the confidence bounds and regions. Now, if $n_h^t(s_h,a_h)>0$, we have
\begin{align*}
\sum_{s'} & \left( q(s'|s_h,a_h) - p_h(s'|s_h,a_h) \right) U_{h+1}^t(s',\pi_{h+1}^{t+1}(s'))\\
 &\leq \lVert q(\cdot|s_h,a_h) - p_h(\cdot|s_h,a_h) \rVert_1 \; \lVert U_{h+1}^t(\cdot,\pi_{h+1}^{t+1}(\cdot))\rVert_\infty\\
 &\leq \sigma_{H-h} \left( \lVert q(\cdot|s_h,a_h) - \hat{p}_h^t(\cdot|s_h,a_h) \rVert_1 + \lVert p_h(\cdot|s_h,a_h) - \hat{p}_h^t(\cdot|s_h,a_h) \rVert_1 \right)\\
 &\leq \sigma_{H-h} \left( \sqrt{ 2 \KL\!\left(\hat{p}_h^t(\cdot|s_h,a_h),q(\cdot|s_h,a_h)\right) } + \sqrt{ 2 \KL\!\left(\hat{p}_h^t(\cdot|s_h,a_h),p_h(\cdot|s_h,a_h)\right) } \right)\\
 &\leq 2\sqrt{2}\sigma_{H-h} \sqrt{ \frac{ \beta( n_h^t(s_h,a_h), \delta) } {n_h^t(s_h,a_h) \vee 1} },
\end{align*}
where we have used Pinsker's inequality to bound the $L^1$-norm using the KL divergence, combined with the fact that both $q$ and $p$ are close to the empirical transition probabilities $\hat{p}^t$ under $\cE$.
\end{proof}

\noindent
As a consequence, we can express the upper bound $U^t$ in terms of the true transition probabilities $p$.

\begin{corollary}\label{cor:upperbound}
If $\cE$ holds, for each $h\in[H]$ and each $(s_h,a_h)$,
\[
U_h^t(s_h,a_h) \leq u_h^t(s_h,a_h) + \gamma\sum_{s'}p_h(s'|s_h,a_h) U_{h+1}^t(s',\pi_{h+1}^{t+1}(s')) + 2\sqrt{2}\gamma\sigma_{H-h} \sqrt{ \frac{ \beta( n_h^t(s_h,a_h), \delta) } {n_h^t(s_h,a_h) \vee 1} }.
\]
\end{corollary}

\noindent
We can also express the lower bound $L^t$ in terms of the transition probabilities $p$ and policy $\pi^{t+1}$.

\begin{lemma}\label{lemma:lowerbound}
If $\cE$ holds, for each $h\in[H]$ and each $(s_h,a_h)$,
\[
L_h^t(s_h,a_h) \geq \ell_h^t(s_h,a_h) + \gamma\sum_{s'}p_h(s'|s_h,a_h) L_{h+1}^t(s',\pi_{h+1}^{t+1}(s')) - 2\sqrt{2}\gamma\sigma_{H-h} \sqrt{ \frac{ \beta( n_h^t(s_h,a_h), \delta) } {n_h^t(s_h,a_h) \vee 1} }.
\]
\end{lemma}

\begin{proof}
We exploit the fact that for each $h\in[H]$, each $(s_h,a_h)$ and each $q \in \cC_h^t(s_h,a_h)$,
\[
\sum_{s'} \left( q(s'|s_h,a_h) - p_h(s'|s_h,a_h) \right) \max_{a'} L_{h+1}^t(s',a') \geq - 2\sqrt{2}\sigma_{H-h} \sqrt{ \frac{ \beta( n_h^t(s_h,a_h), \delta) } {n_h^t(s_h,a_h) \vee 1} }.
\]
The proof is analogous to the proof of Lemma~\ref{lemma:KLball}. We can now write
\begin{align*}
L_h^t&(s_h,a_h) = \ell_h^t(s_h,a_h) + \gamma \min_{p \in \cC_h^t(s_h,a_h)} \sum_{s'} p(s'|s_h,a_h) \max_{a'} L_{h+1}^t(s',a')\\
 &\geq \ell_h^t(s_h,a_h) + \gamma \sum_{s'} p_h(s'|s_h,a_h) \max_{a'} L_{h+1}^t(s',a') - 2\sqrt{2}\gamma\sigma_{H-h} \sqrt{ \frac{ \beta( n_h^t(s_h,a_h), \delta) } {n_h^t(s_h,a_h) \vee 1} }\\
 &\geq \ell_h^t(s_h,a_h) + \gamma \sum_{s'} p_h(s'|s_h,a_h) L_{h+1}^t(s',\pi_{h+1}^{t+1}(s')) - 2\sqrt{2}\gamma\sigma_{H-h} \sqrt{ \frac{ \beta( n_h^t(s_h,a_h), \delta) } {n_h^t(s_h,a_h) \vee 1} }.
\end{align*}
\end{proof}

\noindent
We are now ready to state Lemma~\ref{lemma:difference}.
\begin{lemma}\label{lemma:difference}
If $\cE$ holds, for all $h\in[H]$, $s_h\in\cS_h(\pi^{t+1})$ and $a_h$,
\[D_h^t(s_h,a_h) \leq  \sigma_{H-h+1} \left[4 \sqrt{2}\sqrt{ \frac {\beta(n_h^t(s_h,a_h), \delta)} {n_h^t(s_h,a_h)} } \wedge 1\right]+ \gamma\sum_{s'} p_h(s'|s_h,a_h) D_{h+1}^t(s',\pi_{h+1}^{t+1}(s')).\]
\end{lemma}

\begin{proof}
The bound on the diameter follows directly from Corollary~\ref{cor:upperbound} and Lemma~\ref{lemma:lowerbound}:
\begin{align*}
D_h^t&(s_h,a_h) = U_h^t(s_h,a_h) - L_h^t(s_h,a_h)\\
 &\leq \left( u_h^t(s_h,a_h) - \ell_h^t(s_h,a_h) \right) + \gamma\sum_{s'} p_h(s'|s_h,a_h) \left( U_{h+1}^t(s',\pi_{h+1}^{t+1}(s')) - L_{h+1}^t(s',\pi_{h+1}^{t+1}(s')) \right)\\
 & \;\;\;\; + 4\sqrt{2}\gamma\sigma_{H-h} \sqrt{ \frac{ \beta( n_h^t(s_h,a_h), \delta) } {n_h^t(s_h,a_h) \vee 1} }\\
 &\leq 4 \sqrt{2} \sigma_{H-h+1} \sqrt{ \frac{ \beta( n_h^t(s_h,a_h), \delta) } {n_h^t(s_h,a_h) \vee 1} } + \gamma\sum_{s'} p_h(s'|s_h,a_h) D_{h+1}^t(s',\pi_{h+1}^{t+1}(s')),
\end{align*}
where we used $\cE^r\supseteq\cE$ and Pinsker's inequality to bound \[u_h^t(s_h,a_h) - \ell_h^t(s_h,a_h)\leq\sqrt{ \frac{ 2\beta( n_h^t(s_h,a_h), \delta) } {n_h^t(s_h,a_h) \vee 1} } < 4\sqrt{2} \sqrt{ \frac{ \beta( n_h^t(s_h,a_h), \delta) } {n_h^t(s_h,a_h) \vee 1} }\;.\]
To obtain the final expression in Lemma~\ref{lemma:difference}, we observe that it also trivially holds that 
\[D_h^{t}(s_h,a_h) \leq \sigma_{H-h+1} \leq \sigma_{H-h+1} + \gamma\sum_{s'} p_h(s'|s_h,a_h) D_{h+1}^t(s',\pi_{h+1}^{t+1}(s'))\;,\]
hence
\[D_h^{t}(s_h,a_h) \leq \sigma_{H-h+1} \min\left[4\sqrt{2}\sqrt{ \frac{ \beta( n_h^t(s_h,a_h), \delta) } {n_h^t(s_h,a_h) \vee 1} }, 1\right] + \gamma\sum_{s'} p_h(s'|s_h,a_h) D_{h+1}^t(s',\pi_{h+1}^{t+1}(s'))\;.\]
The conclusion follows by observing that one can get rid of the maximum with 1 in the denominator by using instead the convention $1/0 = +\infty$.
\end{proof}

\subsection{Relating counts to pseudo-counts}
\label{app:counts}

We now assume that the event $\cE$ holds and fix some $h \in [H]$ and some state-action pair $(s_h,a_h)$. For every $\ell \geq h$, we define $p^{\pi}_{h,\ell}(s,a | s_h,a_h)$ to be the probability that starting from $(s_h,a_h)$ in step $h$ and following $\pi$ thereafter, we end up in $(s,a)$ in step $\ell$. We use $p^{t}_{h,\ell}(s,a |s_h,a_h)$ as a shorthand for $p^{\pi^t}_{h,\ell}(s,a |s_h,a_h)$. 

Introducing the \emph{conditional pseudo-counts} $\bar n_{h,\ell}^{t}(s,a ; s_h,a_h)\eqdef\sum_{i=1}^{t}  p^{i}_{h}(s_h,a_h)p^{i}_{h,\ell}\left(s,a | s_h,a_h\right)$ and using that on the event $\cE^{\cnt} \supseteq \cE$ the counts are close to the pseudo-counts, one can prove:
\begin{lemma}\label{lemma:technicalCounts}
If the event $\cE^{\cnt}$ holds, $\left[\sqrt{ \tfrac {\beta(n_{\ell}^t(s,a), \delta)} {n_{\ell}^t(s,a)} } \wedge 1\right] \leq 2\sqrt{\tfrac{\beta(\overline{n}_{h,\ell}^t(s,a ; s_h,a_h), \delta)}{\overline{n}_{h,\ell}^t(s,a ; s_h,a_h) \vee 1}}$.
\end{lemma}

\begin{proof}
As the event $\cE^{\mathrm{cnt}}$ holds, we know that for all $t < \tau$, 
\begin{eqnarray*}n_{\ell}^{t}(s,a) &\geq& \frac{1}{2}\bar n_{\ell}^{t}(s,a) - \beta^{\cnt}(\delta) \\
 & \geq & \frac{1}{2}\bar n_{h,\ell}^{t}(s,a ; s_h,a_h) - \beta^{\cnt}(\delta).
\end{eqnarray*}
We now distinguish two cases. First, if $\beta^{\cnt}(\delta) \leq \tfrac{1}{4}\bar n_{h,\ell}^{t}(s,a ; s_h,a_h)$, then \[\sqrt{\frac {\beta(n_{\ell}^t(s,a), \delta)} {n_{\ell}^t(s,a)}} \leq \sqrt{\frac {\beta\left(\tfrac{1}{4}\bar n_{h,\ell}^{t}(s,a ; s_h,a_h), \delta\right)} {\tfrac{1}{4}\bar n_{h,\ell}^{t}(s,a ; s_h,a_h)}} \leq 2   \sqrt{\frac {\beta\left(\bar n_{h,\ell}^{t}(s,a ; s_h,a_h), \delta\right)} {\bar n_{h,\ell}^{t}(s,a ; s_h,a_h) \vee 1}},\]
where we use that $x \mapsto \sqrt{\beta(x,\delta)/x}$ is non-increasing for $x\geq 1$,  $x \mapsto \beta(x,\delta)$ is non-decreasing, and $\beta^{\cnt}(\delta) \geq 1$. If $\beta^{\cnt}(\delta) > \tfrac{1}{4}\bar n_{h,\ell}^{t}(s,a ; s_h,a_h)$, simple algebra shows that  
\[1 < 2 \sqrt{\frac{\beta^{\cnt}(\delta)}{\overline{n}_{h,\ell}^t(s,a ; s_h,a_h) \vee 1}} \leq 2 \sqrt{\frac{\beta(\overline{n}_{h,\ell}^t(s,a ; s_h,a_h), \delta)}{\overline{n}_{h,\ell}^t(s,a ; s_h,a_h) \vee 1}},\]
where we use that $\beta^{\cnt}(\delta)\leq \beta(0,\delta)$ and $x \mapsto \beta(x,\delta)$ is non-decreasing. If $\overline{n}_{h,\ell}^t(s,a ; s_h,a_h)<1$, the expression uses the trivial bound $\beta^{\cnt}(\delta) > \frac 1 4$. In both cases, we have
\[\left[\sqrt{ \frac {\beta(n_{\ell}^t(s,a), \delta)} {n_{\ell}^t(s,a)} } \wedge 1\right] \leq 2 \sqrt{\frac{\beta(\overline{n}_{h,\ell}^t(s,a ; s_h,a_h), \delta)}{\overline{n}_{h,\ell}^t(s,a ; s_h,a_h) \vee 1}}.\]
\end{proof}

\subsection{Detailed proof of Theorem~\ref{thm:newthm}} \label{subsec:detail}
We assume that the event $\cE$ holds and fix some $h \in [H]$ and some state-action pair $(s_h,a_h)$. We  define some notion of expected diameter in a future step $\ell$ given that $(s_h,a_h)$ is visited at step $h$ under policy $\pi^{t+1}$. For every $(h,\ell) \in [H]^2$ such that $h\leq \ell$ we let
\[q^{t}_{h,\ell}(s_h,a_h) \eqdef \sum_{(s,a)} p^{t+1}_{h}(s_h,a_h)p^{t+1}_{h,\ell}\left(s,a | s_h,a_h\right) D^{t}_{\ell}(s,a).\]
To be more accurate, $q^{t}_{h,\ell}(s_h,a_h)$ is equal to the probability that $(s_h,a_h)$ is visited by $\pi^{t+1}$, multiplied by the expected diameter of the state-action pair $(s,a)$ that is reached at step $\ell$ if one applies $\pi^{t+1}$ after choosing $a_h$ in state $s_h$. In particular, $q^{t}_{h,\ell}(s_h,a_h) = 0$ if $a_h \neq \pi^{t+1}(s_h)$.

\paragraph{Step 1: lower bounding $q^{t}_{h,h}(s_h,a_h)$ in terms of the gaps} From the above definition, 
\[q^{t}_{h,h}(s_h,a_h) = p^{t+1}_{h}(s_h,a_h)D_h^t(s_h,a_h).\]
Using Lemma~\ref{lem:allinone} and the fact that $p^{t+1}_{h}(s_h,a_h) = 0$ if $a_h \neq \pi^{t+1}(s_h)$ yields
\begin{equation}\text{if } t< \tau, \ \ q^{t}_{h,h}(s_h,a_h) \geq \frac{1}{2}p^{t+1}_{h}(s_h,a_h)\Delta_h(s_h,a_h).\label{eq:LBq}\end{equation}

\paragraph{Step 2: upper bounding $q^{t}_{h,h}(s_h,a_h)$ in terms of the counts} Using Lemma~\ref{lemma:difference} and the fact that 
\begin{eqnarray*}
&&\sum_{(s,a)} p_h^{t+1}(s_h,a_h) p_{h,\ell}^{t+1}(s,a | s_h,a_h)\left[ \sum_{(s',a')} p_{\ell}(s' |s,a) \ind\left(a' = \pi^{t+1}_{\ell+1}(s')\right) D_{\ell+1}^{t}(s',a')\right] \\
& =& \sum_{(s',a')}  p_h^{t+1}(s_h,a_h) \underbrace{\left[\sum_{(s,a)}  p_{h,\ell}^{t+1}(s,a | s_h,a_h)p_{\ell}(s' | s,a) \ind\left(a' = \pi^{t+1}_{\ell+1}(s')\right)\right]}_{ = p_{h,\ell+1}^{t+1}(s',a' | s_h,a_h)} D_{\ell+1}^{t}(s',a'),
\end{eqnarray*}
one can establish the following relationship between $q^{t}_{h,\ell}(s_h,a_h)$ and $q^{t}_{h,\ell+1}(s_h,a_h)$: 
\[q^{t}_{h,\ell}(s_h,a_h) \leq \sum_{(s,a)} p^{t+1}_{h}(s_h,a_h)p^{t+1}_{h,\ell}\left(s,a | s_h,a_h\right)\left[4 \sqrt{2}\sqrt{ \frac {\beta(n_h^t(s,a), \delta)} {n_h^t(s,a)} } \wedge 1\right] + \gamma q_{h,\ell+1}^{t+1}(s_h,a_h).\]
By induction, one then obtains the following upper bound: {\small
\begin{equation}\label{eq:UBq}
 q^{t}_{h,h}(s_h,a_h) \leq \sum_{\ell = h}^{H} \gamma^{\ell - h}\sigma_{H-\ell+1}\sum_{(s,a)} p^{t+1}_{h}(s_h,a_h)p^{t+1}_{h,\ell}\left(s,a | s_h,a_h\right)\left[4 \sqrt{2}\sqrt{ \frac {\beta(n_{\ell}^t(s,a), \delta)} {n_{\ell}^t(s,a)} } \wedge 1\right].
\end{equation}}

\paragraph{Step 3: summing the inequalities to get an upper bound on $\overline{n}_h^{t}(s_h,a_h)$} Summing for $t \in \{0,\dots,\tau-1\}$ the inequalities given by \eqref{eq:LBq} yields 
\[\sum_{t=0}^{\tau-1} q^t_{h,h} \geq \frac{\tilde \Delta_h(s_h,a_h)}{2} \left(\sum_{t=0}^{\tau-1}p^{t+1}_{h}(s_h,a_h)\right) = \frac{\tilde \Delta_h(s_h,a_h)}{2} \bar n_h^{\tau}(s_h,a_h).\]
Summing the upper bounds in \eqref{eq:UBq} yields that $\tilde \Delta_h(s_h,a_h) n_h^{\tau}(s_h,a_h)$ is upper bounded by 
\[B_{h}^{\tau}(s_h,a_h) \eqdef 2\sum_{t=0}^{\tau-1}\sum_{\ell = h}^{H} \gamma^{\ell - h}\sigma_{H-\ell+1}\sum_{(s,a)} p^{t+1}_{h}(s_h,a_h)p^{t+1}_{h,\ell}\left(s,a | s_h,a_h\right)\left[4 \sqrt{2}\sqrt{ \frac {\beta(n_{\ell}^t(s,a), \delta)} {n_{\ell}^t(s,a)} } \wedge 1\right]. \]
The rest of the proof consists in upper bounding $B_{h}^{\tau}(s_h,a_h)$ in terms of the pseudo counts $\overline{n}_h^{\tau}(s_h,a_h)$.

\paragraph{Step 4: from counts to pseudo-counts}For all $\ell \geq h$, we introduce the set $\cS_{\ell}(s_h,a_h)$ of states-action pairs $(s,a)$ that can be reached at step $\ell$ from $(s,a)$. 

For each  $(s,a) \in \cS_{\ell}(s_h,a_h)$, we define
\[C_{\ell}(s,a ; s_h,a_h) = \sum_{t=0}^{\tau-1} p^{t+1}_{h}(s_h,a_h)p^{t+1}_{h,\ell}\left(s,a | s_h,a_h\right)\left[4 \sqrt{2}\sqrt{ \frac {\beta(n_{\ell}^t(s,a), \delta)} {n_{\ell}^t(s,a)} } \wedge 1\right].\]
One can observe that $B_{h}^{\tau}(s_h,a_h) =  2\sum_{\ell = h}^{H} \sum_{(s,a) \in \cS_{\ell}(s_h,a_h)}\gamma^{\ell - h}\sigma_{H-\ell+1} C_{\ell}(s,a ; s_h,a_h)$. To upper bound $C_{\ell}(s,a ; s_h,a_h)$ we further introduce the \emph{conditional pseudo-counts} 
\[\bar n_{h,\ell}^{t}(s,a ; s_h,a_h) \eqdef \sum_{i=1}^{t} p^{i}_{h}(s_h,a_h)p^{i}_{h,\ell}(s,a | s_h,a_h),\]
for which one can write 
\[C_{\ell}(s,a ; s_h,a_h) = \sum_{t=0}^{\tau-1}[\bar n_{h,\ell}^{t+1}(s,a ; s_h,a_h) - \bar n_{h,\ell}^{t}(s,a ; s_h,a_h)]\left[4 \sqrt{2}\sqrt{ \frac {\beta(n_{\ell}^t(s,a), \delta)} {n_{\ell}^t(s,a)} } \wedge 1\right].\]
Using Lemma~\ref{lemma:technicalCounts} to relate the counts to the conditional pseudo-counts, one can write
\begin{align*}
 C_{\ell}(s,a ; s_h,a_h)  &\leq 8\sqrt{2}\sum_{t=0}^{\tau -1}[\bar n_{h,\ell}^{t+1}(s,a ; s_h,a_h) - \bar n_{h,\ell}^{t}(s,a ; s_h,a_h)] \sqrt{\frac{\beta(\overline{n}_{h,\ell}^t(s,a ; s_h,a_h), \delta)}{\overline{n}_{h,\ell}^t(s,a ; s_h,a_h) \vee 1}} \\
& \leq  8\sqrt{2}\sqrt{\beta(\overline{n}_{h,\ell}^{\tau}(s,a ; s_h,a_h), \delta)}\sum_{t=0}^{\tau -1}\frac{\bar n_{h,\ell}^{t+1}(s,a ; s_h,a_h) - \bar n_{h,\ell}^{t}(s,a ; s_h,a_h)}{\sqrt{\overline{n}_{h,\ell}^t(s,a ; s_h,a_h) \vee 1}} \\
& \leq 8\sqrt{2}(1+\sqrt{2}) \sqrt{\beta(\overline{n}_{h,\ell}^{\tau}(s,a ; s_h,a_h), \delta) \times \bar n^\tau_{h,\ell}(s,a ; s_h,a_h)},
\end{align*}
where the last step uses Lemma 19 in \cite{UCRL10}. 

Finally,  by summing over episodes $\ell$ and over reachable states $(s,a) \in\cS_{\ell}(s_h,a_h)$, we can upper bound $B^{\tau}_h(s_h,a_h)$ by {\small
\begin{align*}
&2 \sum_{\ell =h}^{H} \gamma^{\ell - h}\sigma_{H-\ell+1} \left[ 8\sqrt{2}(1+\sqrt{2})\sqrt{\beta(\overline{n}_{h}^{\tau}(s_h,a_h), \delta)}\sum_{(s,a) \in \cS_{\ell}(s_h,a_h)}  \sqrt{\bar n^\tau_{h,\ell}(s,a ; s_h,a_h)}\right] \\
& \leq 2 \sum_{\ell =h}^{H} \gamma^{\ell - h}\sigma_{H-\ell+1} \left[ 8\sqrt{2}(1+\sqrt{2})\sqrt{\beta(\overline{n}_{h}^{\tau}(s_h,a_h), \delta)}\sqrt{ (BK)^{h-\ell}} \sqrt{\sum_{(s,a) \in \cS_{\ell}(s_h,a_h)}\bar n^\tau_{h,\ell}(s,a ; s_h,a_h)}\right] \\
& = 2 \sum_{\ell =h}^{H} \gamma^{\ell - h}\sigma_{H-\ell+1} \left[ 8\sqrt{2}(1+\sqrt{2})\sqrt{\beta(\overline{n}_{h}^{\tau}(s_h,a_h), \delta)}\sqrt{ (BK)^{h-\ell}} \sqrt{\bar n^\tau_{h}(s_h,a_h)}\right], 
\end{align*} }
where we have used that $\sum_{(s,a) \in \cS_{\ell}(s_h,a_h)}\bar n^\tau_{h,\ell}(s,a ; s_h,a_h) = \bar n_h^{\tau}(s_h,a_h)$. By using further Lemma~\ref{lemma:constants} to upper bound all the constants, we obtain 
\[B^{\tau}_h(s_h,a_h) \leq 64\sqrt{2}(1+\sqrt{2}) \left(\sqrt{BK}\right)^{H-h} \sqrt{\bar n^\tau_{h}(s_h,a_h)\beta(\overline{n}_{h}^{\tau}(s_h,a_h), \delta)}\;.\]

\begin{lemma}
\label{lemma:constants} For every $x > 1$, \ $\sum_{\ell=h}^{H} (\gamma x)^{\ell-h}\sigma_{H-\ell+1} \leq \frac{x^{H-h}}{\left(1-\frac{1}{x}\right)^2}$.
\end{lemma}

\begin{proof}
Since $\gamma \leq 1$ and $x > 1$, we can write
\begin{align*}
\sum_{\ell=h}^{H} (\gamma x)^{\ell-h}\sigma_{H-\ell+1} &\leq \sum_{\ell=h}^{H} x^{\ell-h}(H-\ell+1) = \sum_{\ell=0}^{H-h} x^{\ell}(H-h-\ell+1)\\
 &= x ^ {H-h} \sum_{\ell=0}^{H-h} \frac {H-h-\ell+1} { x^{H-h-\ell} } = x ^ {H-h} \sum_{\ell=0}^{H-h} (\ell+1) r^\ell,
\end{align*}
where $r=1/x<1$. The latter is an {\em arithmetico-geometric sum} that can be upper bounded as
\begin{align*}
\sum_{\ell=0}^{H-h} (\ell+1) r^\ell & \leq  \sum_{\ell=0}^{\infty} (\ell+1) r^\ell = \frac 1 {(1-r)^2} =  \frac 1 {\left(1 - \frac 1 x\right)^2}\;.
\end{align*}
\end{proof}

\section{Proof of Theorem~\ref{thm:deterministic}} \label{app:deterministic}

The proof of Theorem~\ref{thm:deterministic} uses the same ingredients as the proof of Theorem~\ref{thm:newthm}: Lemma~\ref{lem:allinone} which relates the gaps to the diameters of the confidence intervals $D_h^t(s_h,a_h) = U_h^t(s_h,a_h) - L_h^t(s_h,a_h)$ and a counterpart of Lemma~\ref{lemma:difference} for the deterministic case, stated below. 

\begin{lemma}\label{lemma:difference_deterministic}
If $\cE$ holds, and $t_{1:H} = (s_1,a_1,\dots,s_H,a_H)$ is the $(t+1)$-st trajectory generated by \OurAlgorithm{}, for all $h \in [H]$, 
\[D_h^t(s_h,a_h) \leq \left[\sqrt{\frac{2\beta(n^t_h(s_h,a_h),\delta)}{n^t_h(s_h,a_h)}} \wedge 1 \right] + \gamma D_{h+1}^t(s_{h+1}, a_{h+1}).\]
\end{lemma}

It follows from Lemma~\ref{lemma:difference_deterministic} that for all $h \in [H]$, along the $(t+1)$-st trajectory $t_{1:H} = (s_1,a_1,\dots,s_H,a_H)$, 
\[D_h^t(s_h,a_h) \leq \sum_{\ell = h}^{H}\gamma^{\ell - h}\left[\sqrt{\frac{2\beta(n^t_{\ell}(s_{\ell},a_{\ell}),\delta)}{n^t_{\ell}(s_{\ell},a_{\ell})}} \wedge 1 \right].\]
Letting $n^t(t_{1:H})$ be the number of times the trajectory $t_{1:H}$ has been selected by \OurAlgorithm{} in the first $t$ episodes, one has $n^t_{\ell}(s_{\ell},a_{\ell}) \geq n^t(t_{1:H})$. Hence, if $n^t(t_{1:H}) > 0$, it holds that
\[D_h^t(s_h,a_h) \leq \sum_{\ell = h}^{H}\gamma^{\ell - h}\sqrt{\frac{2\beta(n^t(t_{1:H}),\delta)}{n^t(t_{1:H})}} = \sigma_{H-h +1}\sqrt{\frac{2\beta(n^t(t_{1:H}),\delta)}{n^t(t_{1:H})}} .\]
Using Lemma~\ref{lem:allinone}, if $t < \tau$, if $t_{1:H}$ is the trajectory selected at time $(t+1)$, either $n^t(t_{1:H}) = 0$ or  
\[\forall h \in [H], \ \ \tilde{\Delta}_h(s_h,a_h) \leq \sigma_{H-h +1}\sqrt{\frac{2\beta(n^t(t_{1:H}),\delta)}{n^t(t_{1:H})}} \]
It follows that for any trajectory $t_{1:H}$, 
\[n^{\tau}(t_{1:H}) \left[\max_{h \in [H]} \frac{\left(\tilde{\Delta}_h(s_h,a_h)\right)^2}{(\sigma_{H-h+1})^2}\right] \leq 2\beta(n^{\tau}(t_{1:H}),\delta).\]
The conclusion follows from Lemma~\ref{lemma:technical} and from the fact that $\tau = \sum_{t_{1:H} \in \cT} n^{\tau}(t_{1:H})$.

\section{Sample complexity of Sparse Sampling in the Fixed-Confidence Setting} \label{proof:SS}

In this section, we prove Lemma~\ref{lemma:SS}.

	For simplicity, and without loss of generality, assume that the reward function is known. Let $C > 0$. Sparse Sampling builds, recursively, the estimates $\widehat{V}_h$ and $\widehat{Q}_h$ for $h \in [H+1]$, starting from $\widehat{V}_{H+1}(s) = 0$ and $\widehat{Q}_{H+1}(s, a)=0$ for all $(s, a)$. Then, from a target state-action pair $(s, a)$, it samples $C$ transitions $Z_i \sim p_h(\cdot|s,a)$ for $i \in [C]$ and computes:
	\begin{align*}
		\widehat{Q}_h(s, a) = r_h(s, a) + \frac{1}{C}\sum_{i=1}^C \widehat{V}_{h+1}(Z_i), \; \mbox{ with } 	\widehat{V}_{h}(s) = \max_a \widehat{Q}_h(s, a)
	\end{align*}
	For an initial state $s$, its output is $\widehat{Q}_1(s, a)$ for all $a \in [K]$. For any state $s$, consider the events
	\begin{align*}
		\cG(s, a, h) = \cpa{ \abs{ \widehat{Q}_h(s, a) - Q_h^\star(s, a) } \leq \epsilon_h } \bigcap \cpa{ \bigcap_{z \in \mathrm{supp}\spa{p_h(\cdot|s,a)}  } \cG(z, h+1) }.
	\end{align*}
	and
	\begin{align*}
		\cG(s, h) = \bigcap_{a \in [K]} 	\cG(s, a, h).
	\end{align*}
	defined for $h \in [H+1]$, where $\epsilon_h \eqdef (H-h+1)H\sqrt{(2/C)\log(2/\delta')}$ for some $\delta' > 0$.

	Let
	\begin{align*}
		\delta_h = \frac{2K\delta'}{BK-1} \pa{ (BK)^{H-h+1} - 1 }
	\end{align*}

	We prove that, for all $s$ and all $h$, $\bP\spa{ \cG(s, h) } \geq 1 - \delta_h$. We proceed by induction on $h$. For $h = H + 1$, we have $\widehat{Q}_{H+1}(s, a) =  Q_{H+1}^\star(s, a) = 0$ for all $(s, a)$ by definition, which gives us $\bP\spa{\cG(s, a, H+1)} = 1$ and, consequently, $\bP\spa{\cG(s, H+1)} = 1$.

	Now, assume that $\bP\spa{ \cG(z, h+1) } \geq 1 - \delta_h $ for all $z$. Since
	\begin{align*}
		\abs{\widehat{Q}_h(s, a) - Q^\star_h(s, a)} & \leq \frac{1}{C}\abs{\sum_{i=1}^C  \pa{ \widehat{V}_{h+1}(Z_i) - V_{h+1}^\star(Z_i) } }  + \frac{1}{C}\abs{\sum_{i=1}^C  \pa{  V_{h+1}^\star(Z_i) - \bE\spa{V_{h+1}^\star(Z_i)} } }
	\end{align*}

	We have,

	\begin{align*}
		\bP\spa{ \cG(s, a, h)^\complement } & \leq \sum_{z \in \mathrm{supp}\spa{p_h(\cdot|s,a)}} \bP\spa{ \cG(z, h+1)^\complement } + \bP\spa{ \frac{1}{C}\abs{\sum_{i=1}^C  \pa{  V_{h+1}^\star(Z_i) - \bE\spa{V_{h+1}^\star(Z_i)} } } \geq \epsilon_h - \epsilon_{h+1}  }  \\
		& \leq B\delta_{h+1} + 2\exp\pa{-\frac{ C (\epsilon_h - \epsilon_{h+1})^2 }{ 2H^2 } } \leq B\delta_{h+1} + 2\delta'
	\end{align*}
	and, consequently,
	\begin{align*}
		\bP\spa{ \cG(s, h)^\complement } \leq BK\delta_{h+1} + 2K \delta'= \delta_h.
	\end{align*}
	which gives us $\bP\spa{ \cG(s, h) } \geq 1 - \delta_h$, as claimed above.  In particular, taking $h=1$, we have

	\begin{align*}
		 \abs{\widehat{Q}_1(s, a) - Q_1^\star(s, a)}  \leq  H^2 \sqrt{(2/C)\log(2/\delta')}
	\end{align*}
	with probability at least $1 - \delta$, where $\delta = 2K\delta'\pa{(BK)^H - 1}/(BK-1)$. Finally, we let $\epsilon \eqdef  H^2 \sqrt{(2/C)\log(2/\delta')}/2$ and solve for $C$, obtaining
	\begin{align*}
		C = \BigO{\frac{H^5}{\epsilon^2}\log\pa{\frac{BK}{\delta}}}.
	\end{align*}
	Thus predicting $\hat{a} = \argmax{a}\widehat{Q}_1(s_1, a)$ after $\BigO{C (BK)^H}$  sampled transitions we have
	\[
\bP\left(Q^\star(s_1,\hat{a}_\tau) > Q^\star(s_1,a^\star) - \epsilon\right) \geq 1 - \delta\,.
	\]

\section{A Technical Lemma} \label{proof:technical}

We state and prove below a technical result that permits to obtain an upper bound on $n$ from a condition of the form $n \Delta^2 \leq \beta(n,\delta)$, like the one which appears in Theorem~\ref{thm:newthm}. 

\begin{lemma}
	\label{lemma:technical}
	Let $n \geq 1$ and $a, b, c, d > 0$. If $n \Delta^2 \leq a + b \log( c+ dn)$
	then
	\begin{align*}
	n \leq \frac{1}{\Delta^2}\left[ a + b \log\pa{c + \frac{d}{\Delta^4} (a + b(\sqrt{c} + \sqrt{d}))^2 } \right].
	\end{align*}
\end{lemma}
\begin{proof}
	Since $\log(x) \leq \sqrt{x}$ and $\sqrt{x+y} \leq \sqrt{x}+\sqrt{y}$ for all $x, y > 0$, we have
	\begin{align*}
	& n\Delta^2 \leq a + b \sqrt{c+dn} \leq a + b\sqrt{c} + b\sqrt{d} \sqrt{n} \\
	& \implies \sqrt{n} \Delta^2  \leq \frac{ a + b\sqrt{c}}{\sqrt{n}} +  b\sqrt{d}  \leq a + b(\sqrt{c} + \sqrt{d}) \\
	& \implies n \leq \frac{1}{\Delta^4}\pa{a + b(\sqrt{c} + \sqrt{d})}^2.
	\end{align*}
	Hence,
	\begin{align*}
	& n \Delta^2 \leq a + b \log( c+ dn) \\
	& \implies n \Delta^2 \leq a + b \log( c+ dn)
	\quad \text{ and } \quad n \leq \frac{1}{\Delta^4}\pa{a + b(\sqrt{c} + \sqrt{d})}^2 \\
	& \implies  n \Delta^2 \leq a + b \log\pa{ c+  \frac{d}{\Delta^4}\pa{a + b(\sqrt{c} + \sqrt{d})}^2 }.
	\end{align*}
\end{proof}

\end{document}